\documentclass[10pt,twocolumn,letterpaper]{article}

\usepackage{cvpr}              % To produce the CAMERA-READY version

\usepackage{graphicx}
\usepackage{rotating}
\usepackage{multirow}
\usepackage{amsmath}
\usepackage{amssymb}
\usepackage{booktabs}
\usepackage{float}
\usepackage[switch]{lineno}

\usepackage[pagebackref,breaklinks,colorlinks]{hyperref}
\usepackage[accsupp]{axessibility} % Improves PDF readability for those with disabilities.

\usepackage[capitalize]{cleveref}
\crefname{section}{Sec.}{Secs.}
\Crefname{section}{Section}{Sections}
\Crefname{table}{Table}{Tables}
\crefname{table}{Tab.}{Tabs.}

\def\cvprPaperID{5428} % *** Enter the CVPR Paper ID here
\def\confName{CVPR}
\def\confYear{2022}

\begin{document}
\title{Consistent Explanations by Contrastive Learning}

\author{Vipin Pillai\\
University of Maryland, Baltimore County\\
{\tt\small vp7@umbc.edu}
\and
Soroush Abbasi Koohpayegani\\
University of Maryland, Baltimore County\\
{\tt\small soroush@umbc.edu}
\and
Ashley Ouligian\\
Northrop Grumman\\
{\tt\small Ashley.Rothballer@ngc.com}
\and
Dennis Fong\\
Northrop Grumman\\
{\tt\small Dennis.Fong@ngc.com}
\and
Hamed Pirsiavash\\
University of California, Davis\\
{\tt\small hpirsiav@ucdavis.edu}
}
\maketitle

\begin{abstract}
Post-hoc explanation methods, e.g., Grad-CAM, enable humans to inspect the spatial regions responsible for a particular network decision. However, it is shown that such explanations are not always consistent with human priors, such as consistency across image transformations. Given an interpretation algorithm, e.g., Grad-CAM, we introduce a novel training method to train the model to produce more consistent explanations. Since obtaining the ground truth for a desired model interpretation is not a well-defined task, we adopt ideas from contrastive self-supervised learning, and apply them to the interpretations of the model rather than its embeddings. We show that our method, Contrastive Grad-CAM Consistency (CGC), results in Grad-CAM interpretation heatmaps that are more consistent with human annotations while still achieving comparable classification accuracy. Moreover, our method acts as a regularizer and improves the accuracy on limited-data, fine-grained classification settings. In addition, because our method does not rely on annotations, it allows for the incorporation of unlabeled data into training, which enables better generalization of the model. Our code is available here: \href{https://github.com/UCDvision/CGC}{https://github.com/UCDvision/CGC}
\end{abstract}

\section{Introduction}
Deep neural networks have become ubiquitous in many applications owing to their performance on several computer vision tasks. Although they have been instrumental in achieving state-of-the-art accuracy, deep neural networks are widely considered to be black box systems, which is not desirable. For example, if an AI system is deployed to identify a malignant tumor from CT scans, it is important for medical experts to understand the reasoning behind the decision-making process \cite{tjoa2020survey}. This not only enables building trust, but also helps identify any spurious correlations that the network may have inadvertently learned to use to make its decision \cite{singh2020don}. In recent years, there have been attempts to open this black box by designing frameworks to explain the network's decision-making process. Post-hoc explanation methods such as CAM \cite{Zhou2015LearningDF}, Grad-CAM \cite{Selvaraju2016GradCAMVE}, and Full-Grad \cite{srinivas2019fullgrad} generate a heatmap in the size of the image with higher values corresponding to the regions that contributed most to the network's decision. %Such saliency-based methods enable validation of the resulting heatmaps against human priors.\par

\begin{figure}[t]
\centering
\setlength{\tabcolsep}{1pt}
\begin{tabular}{c c c c c c | c c c c c c}
\hline
 \scriptsize{Original} & \scriptsize{Augmented} & & & & & & & & &  \scriptsize{Original} & \scriptsize{Augmented}\\
\hline

    \hspace{0.5em} \includegraphics[width=.1\textwidth, height=.1\textwidth]{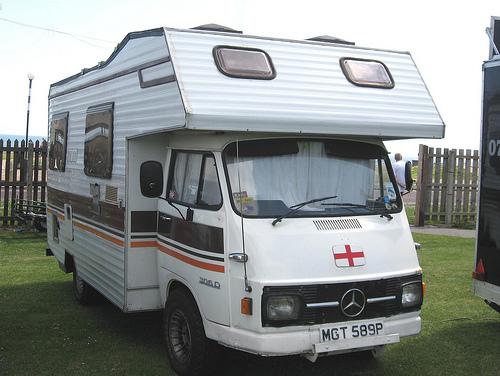}&
    \includegraphics[width=.1\textwidth]{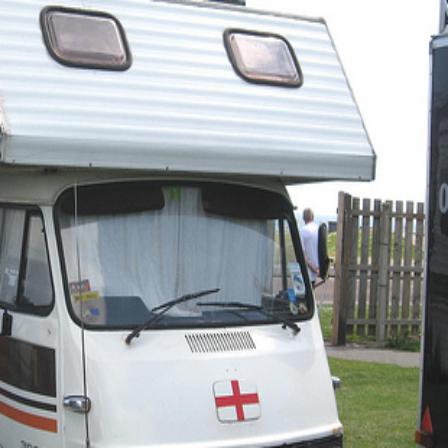} & & & & & & & & &
    \includegraphics[width=.1\textwidth, height=.1\textwidth]{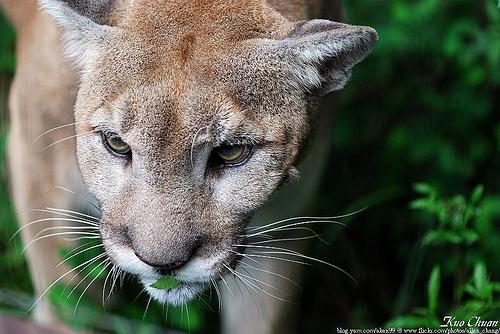}&
    \includegraphics[width=.1\textwidth]{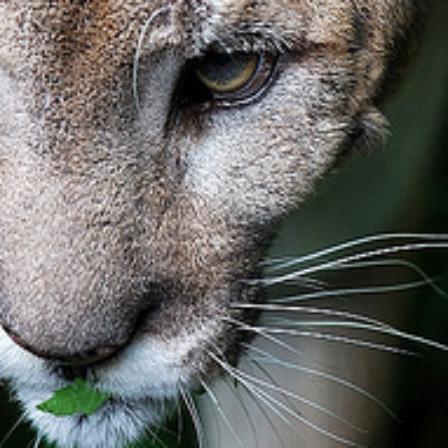}\\
    
    \begin{sideways} \quad \footnotesize{Baseline} \end{sideways} 
    \includegraphics[width=.1\textwidth]{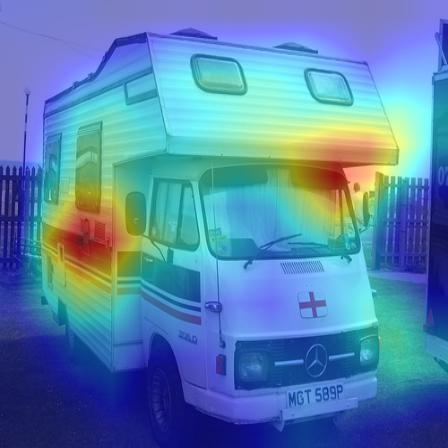}&
    \includegraphics[width=.1\textwidth]{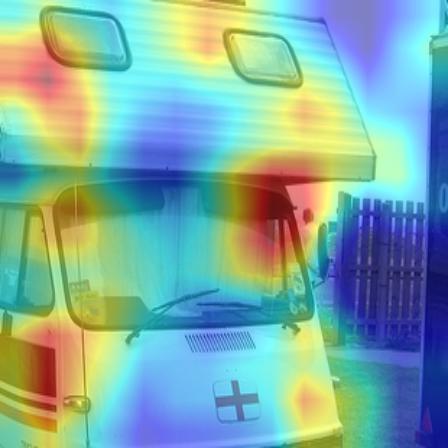} & & & & & & & & &
    \includegraphics[width=.1\textwidth]{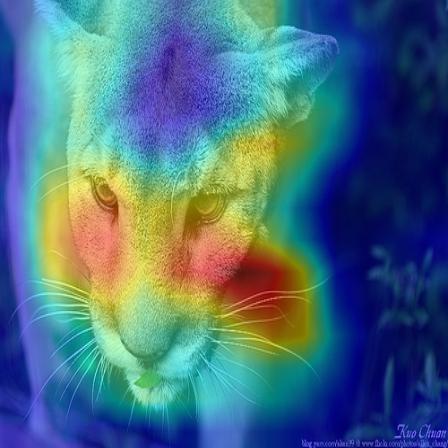}&
    \includegraphics[width=.1\textwidth]{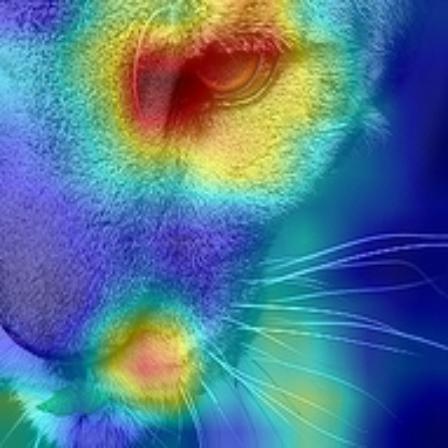}\\
    
    \begin{sideways} \quad  \footnotesize{Ours} \end{sideways} 
    \includegraphics[width=.1\textwidth]{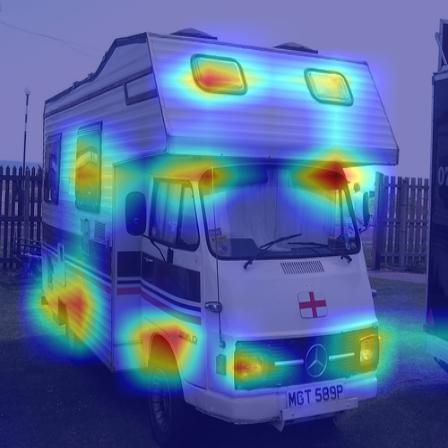}&
    \includegraphics[width=.1\textwidth]{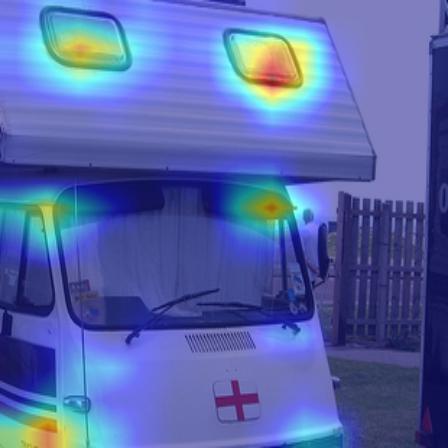} & & & & & & & & &
    \includegraphics[width=.1\textwidth]{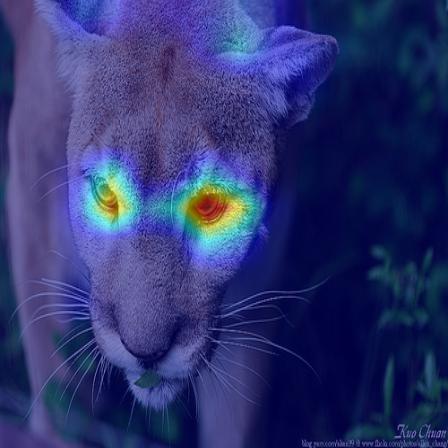}&
    \includegraphics[width=.1\textwidth]{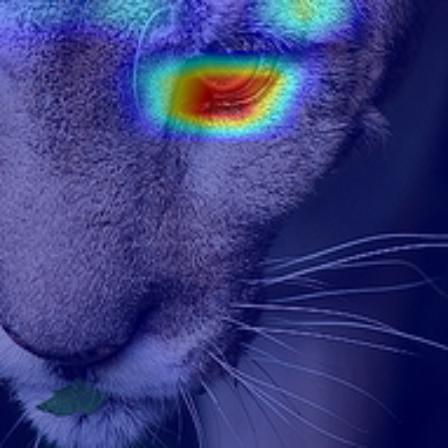}\\
    % \vspace{-0.05in}
    \multicolumn{6}{c}{R.V} & \multicolumn{6}{c}{Puma}\\ 
\end{tabular}
\caption{Our method significantly improves the consistency of Grad-CAM explanation heatmaps under data augmentation. For both the RV and the Puma, our method highlights the same portions of the image in both the original and augmented versions.}
\label{fig:intro_results_cgc}
\end{figure}

\begin{figure*}
\begin{center}
   \includegraphics[width=0.77\linewidth]{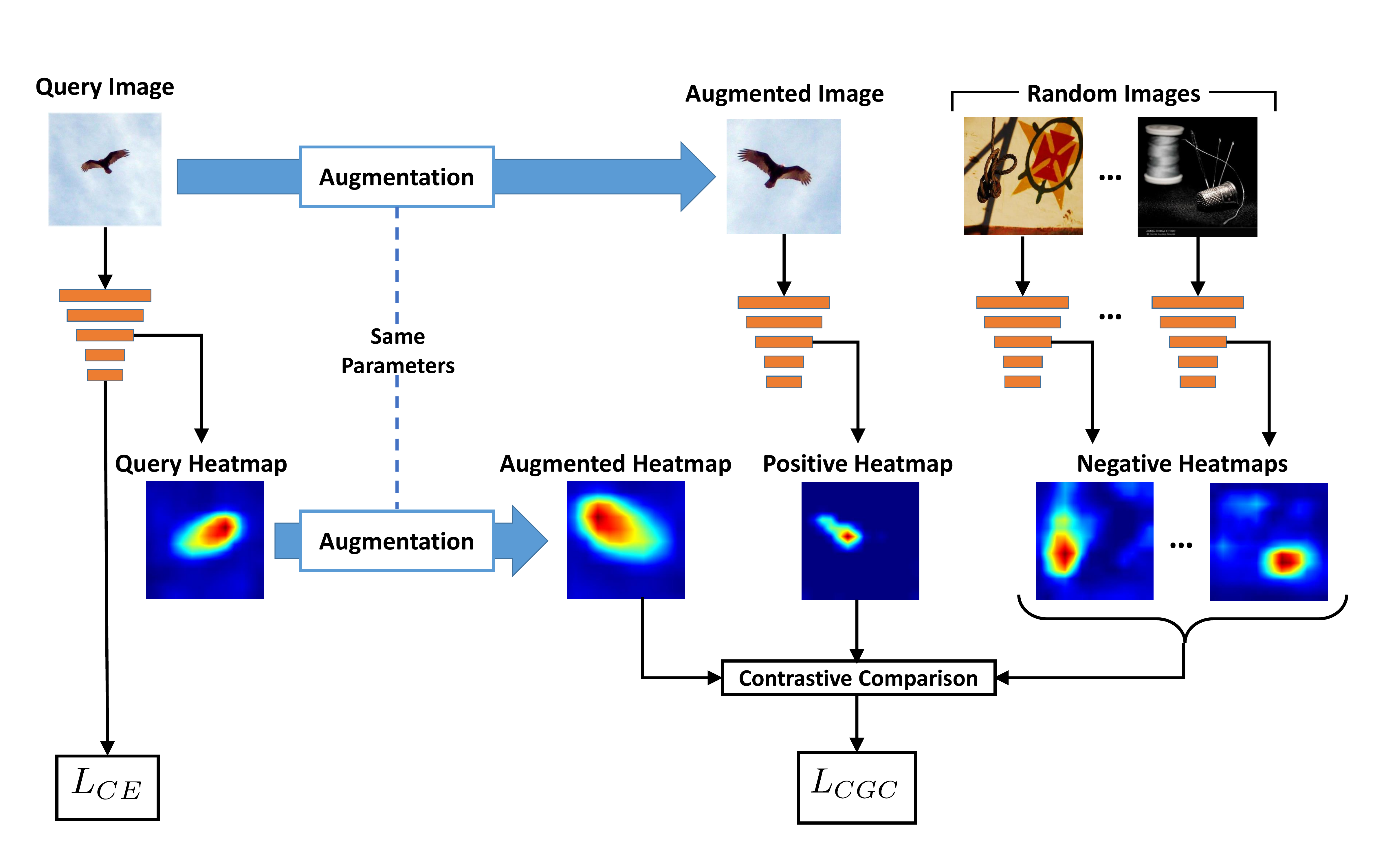}
   \vspace{-.15in}
   \caption{\textbf{The block diagram of our method.} Our method consists of both cross-entropy loss ($L_{CE}$) and contrastive Grad-CAM consistency loss ($L_{CGC}$). We load a batch of random images, and consider one to be the query image. We feed the query image to the network and calculate $L_{CE}$. We calculate Grad-CAM for this image on the top predicted category and then augment the heatmap. We then augment all the images in the batch, feed them to our model, and calculate the Grad-CAM heatmap for the top predicted category. The top category is chosen using the original image and not the augmented one. (Note that all random images are also augmented with independently sampled parameters. We do not show this to reduce the clutter.) The heatmap from the augmented query image is considered to be the positive example. The heatmaps from the other random images in the batch are the negative examples. The augmented query heatmap, the positive heatmap, and the negative heatmaps are all fed into our contrastive comparison function to produce our $L_{CGC}$ term.
   }
   \vspace{-0.25in}
\label{fig:teaser}
\end{center}
\end{figure*}

Unlike image labels, there can be multiple valid explanations for a given image category. Furthermore, a valid explanation might not involve the entire object-segmentation area. For example, in order to correctly classify an image of a dog, the network might rely on just the facial features of the dog, or the texture of the fur and the tail, or a combination of both. Each of these are valid explanations, and hence, generating ground truth annotations for explanations is a not well defined task. This makes it difficult to directly supervise the network's explanation during training.

It is shown that most interpretation methods are not consistent with spatial transformation of the images. For instance, shifting an image does not shift the interpretation heatmap in the same way \cite{Guo_2019_CVPR, kindermans2017reliability}. In addition, in fine-grained visual categorization, it is important to learn the subtle yet discriminative features across classes (e.g., wing color, beak and eyes for a bird) \cite{welinder2010caltech} and hence the network interpretation should focus on the most salient features that discriminate the correct class from other classes. Assuming Grad-CAM is a truthful interpretation method, we are interested in improving the training process of the deep network so that its Grad-CAM interpretation is more consistent with respect to spatial transformations, thus making the model more interpretable. We use Grad-CAM as the key interpretation method for the rest of the paper since it passes the sanity check introduced in \cite{adebayo2018sanity} and is end-to-end differentiable.

% Throughout this paper, we will use Grad-CAM visualization as our interpretation method. 
Inspired by self-supervised learning, we argue that a spatial affine transformation on an image should correspond to such transformation in the interpretation. For instance, given image of a ``dog'', the heatmap for ``dog'' category should highlight the dog and it should shift or zoom if we shift or zoom into the image. Figure \ref{fig:intro_results_cgc} shows an example where the change in Grad-CAM heatmap for the baseline network computed for the ``RV'' and ``Puma'' categories is not consistent with the spatial affine transformation applied to the images. We adopt ideas from recently developed contrastive self-supervised learning literature \cite{Hadsell2006DimensionalityRB, Noroozi2017RepresentationLB, He2019MomentumCF} and design a loss function that encourages the Grad-CAM of an image to be close to the Grad-CAM of an augmented version of the same image while being far from the Grad-CAM of other random images.

We evaluate our CGC method for both classification accuracy and quality of explanations measured by the ``Content Heatmap'' (CH) metric introduced in \cite{pillai2021explainable}. A similar metric was also introduced in \cite{Subramanya_2019_ICCV} to measure the energy of explanation heatmap inside an adversarial patch. CH measures the cumulative heatmap contained within an annotated object mask and is a proxy to measure the consistency of an explanation heatmap with respect to human annotations. Moreover, since our method guides the model to focus mainly on the most discriminative features of the images, it improves the accuracy in fine-grained classification settings while improving the consistency of interpretations. The effect is even more prominent in learning with fewer labels on fine-grained tasks. We believe this is due to the regularization effect that our method adds to the training process. In addition, since our loss function does not need labeled data, it can benefit from unlabeled data during training. Our experiments show that our method improves the consistency of interpretation with respect to human annotations as well as the classification accuracy in limited labels and fine-grained settings.

% Our paper makes the following key contributions:
% \begin{itemize}
%   \item Our CGC method results in classification accuracy comparable to that of a network trained with cross-entropy loss alone while significantly improving the explanation heatmaps.
%   \item We improve the classification accuracy on fine-grained datasets such as Caltech-Birds \cite{welinder2010caltech}, Stanford Cars \cite{KrauseStarkDengFei-Fei_3DRR2013}, VGG Flowers \cite{Nilsback08}, and FGVC Aircraft \cite{maji13fine-grained} datasets in both full- and limited-data settings.
%   \item We incorporate unlabeled data into our training method for our contrastive loss term and show improved classification accuracy over the baseline trained with only the labeled data.
% \end{itemize}

\section{Related Work}
\noindent \textbf{Explanation Methods:}
Early methods used to explain the decision-making process of a deep neural network have relied on developing post-hoc explanation frameworks that generate explanation heatmaps for a given network, input image, and the corresponding predicted class. \cite{Simonyan2013DeepIC} introduced a method that measured the gradient of the predicted class with respect to the input image and used spatial locations with large gradient magnitudes to obtain a saliency map. This was improved upon by \cite{Springenberg2014StrivingFS,  Sundararajan2017AxiomaticAF} to obtain sharper saliency maps. Such gradient-based saliency methods have been shown to match the state-of-the-art example-based explanations on several datasets \cite{ijcai2017-371}. Class Activation Mapping (CAM) \cite{Zhou2015LearningDF} was introduced to generate a coarse localization heatmap by using a global average pooling (GAP) layer to compute the gradients flowing into the final convolution layer. Gradient-weighted Class Activation Mapping (Grad-CAM) \cite{Selvaraju2016GradCAMVE} generalized CAM by eliminating the need for a GAP layer to compute coarse localization heatmaps. Another class of explanation methods perturb the input to the model and observe the resulting changes to the output \cite{Zeiler2013VisualizingAU, NIPS2017_0060ef47, fong2017interpretable, Petsiuk2018rise, fong2019understanding}.

\noindent \textbf{Self-Supervised Learning:} Self-supervised learning methods rely on introducing pretext tasks based on structural priors of the data without using image labels to learn meaningful representations. 
Spatial structure in the visual domain \cite{Doersch2015UnsupervisedVR, norooziECCV16, Noroozi2017RepresentationLB}, color information \cite{Zhang2016ColorfulIC, Larsson2017ColorizationAA}, and spatial orientation \cite{gidaris2018unsupervised} have been used to design pretext tasks.
% \cite{Doersch2015UnsupervisedVR, norooziECCV16, Noroozi2017RepresentationLB} relied on spatial structure in the visual domain to design pretext tasks. Additional pretext tasks have been proposed that rely on color information \cite{Zhang2016ColorfulIC, Larsson2017ColorizationAA} and spatial orientation \cite{gidaris2018unsupervised}.
Recently, such hand-crafted pretext tasks have been superseded by methods that instead learn representations by contrasting the feature vectors of positive pairs with those of negative pairs \cite{He2019MomentumCF, chen2019looks}. We build upon these ideas by leveraging contrastive learning on the network's explanation heatmaps instead of the representations. This aligns with our prior that the network's explanation for a given image should be consistent under augmentations of that image. 

\noindent \textbf{Consistency for Explanations:} \cite{rieger2020interpretations} use domain knowledge to align explanations with prior knowledge in the form of importance scores. 
\cite{han2021explanation} use adversarial perturbation on the input images for explanation consistency with the original image.  
\cite{wang2020proactive} use causal masking to remove salient regions of the input image and generate positive and negative contrast images to improve model interpretability.
\cite{jacovi-etal-2021-contrastive,jha2021supervised} propose contrastive learning to improve interpretability for NLP models.
\cite{Guo_2019_CVPR} introduced the idea of imposing a perceptual consistency prior on the attention heatmaps while training the network for multi-label image classification. The key idea in \cite{Guo_2019_CVPR} is that the CAM \cite{Zhou2015LearningDF} attention heatmap of an image should follow the same transformation if the image is transformed. A similar idea to enforce consistency regularization on the CAM attention heatmaps using the concept of attention separability and cross-layer attention consistency \cite{Wang_2019_ICCV} was introduced for the task of weakly-supervised semantic segmentation.

Our work differs from these in that we use a set of negative examples along with the standard image transformations in a contrastive setting. Negative examples play an important role in ensuring that the interpretation relies on image-specific regions rather than being biased towards a blob in the middle of the image or spread around the image uniformly. Moreover, most of these works evaluate the interpretation by using it as a semantic segmentation tool. However, we believe that an interpretation heatmap should not necessarily highlight the whole object mask. Instead, it should highlight the ``most discriminative regions'' of an image, which is a strict subset of the semantic segmentation mask. We emphasize that we are not introducing a new explanation method, but rather learning a model that is tuned to be explainable for a given explanation method.

Probably, \cite{pillai2021explainable} is the closest to our work as it uses self-supervised learning ideas to remove spurious correlations in the interpretation heatmap. \cite{pillai2021explainable} uses a different self-supervised pseudo task for reducing the contextual reasoning by feeding synthetic composite images during training which is out-of-distribution data. Our work is focused on contrastive learning which has driven the recent progress in self-supervised learning community and uses in-distribution images only during training. %We show that our method outperforms \cite{pillai2021explainable}
Unlike \cite{pillai2021explainable}, our method does not require labels to encourage explanation consistency and hence we are able to leverage additional unlabeled data for our contrastive explanation consistency loss together with the labeled data used for the standard cross-entropy loss.
% and we show improved results using just 1\% labeled data on the ImageNet dataset. 
% Moreover, our loss uses unlabeled natural images instead of labeled synthetic composite ones and achieves better results compared to \cite{pillai2021explainable}. %Also, we show good results when using unlabeled data as negatives.

%We use random unlabeled negative images instead of distractor images. while our method uses natural images along with contrastive learning. We compare with \cite{pillai2021explainable} in our experiments.

\section{Method}

Figure \ref{fig:teaser} shows a block diagram of our method. Our training process consists of both categorical cross-entropy loss ($L_{CE}$) and contrastive Grad-CAM consistency loss ($L_{CGC}$). In this section, we first present a brief overview of the Grad-CAM interpretation algorithm and then describe the contrastive Grad-CAM consistency loss term.

\noindent {\bf Background on Grad-CAM \cite{Selvaraju2016GradCAMVE}:}
Consider an input image $x$ and a deep neural network $f$. Let $y$ be the vector of output logits when we feed $x$ to the model $f$ where, $y^t$ corresponds to the output for category $t$. The Grad-CAM of  model $f$ for image $x$ and a given category $t$ is a heatmap that highlights the regions of image $x$ responsible for the model's classification of the image as category $t$. We calculate this heatmap by choosing an intermediate convolutional layer and then linearizing the rest of the network to be interpretable. More specifically, we calculate the derivative of the predicted output with respect to each channel of the convolutional layer averaged over all spatial locations. This results in a scalar for each channel that captures the importance of that channel in making the current prediction. Then, we calculate a weighted average of all activations of the convolutional layer with the above importance weights for each channel to get a 2D matrix over spatial locations. Finally, we keep only positive numbers and resize it to the size of input image to get the interpretation heatmap.

\subsection{Contrastive Grad-CAM Consistency Loss}
We are interested in training the image classification model so that its Grad-CAM heatmaps are consistent with spatial transformations. For instance, when we shift the image, the interpretation heatmap should also shift in the same way. Also, the heatmap should be specific to the image, as in not always focused on a blob in the middle of the image or be spread around the whole image.

Inspired by contrastive learning methods in self-supervised learning, we design a contrastive loss function for the interpretation heatmaps that acts as a regularizer when added to the standard cross entropy loss for supervised learning. We want the transformed interpretation of a query image to be close to the interpretation of the transformed query image while being far from interpretations of other random images.

More formally, we assume $g(.)$ is the Grad-CAM operator that calculates the interpretation heatmap for the top predicted category of an input image. Given a set of $n$ random images $\{x_j\}_{j=1}^n$, we augment them with independent random spatial transformations $T_j(.)$ which involves a combination of random scaling, cropping, and flipping. This is similar to the standard augmentation usually done in deep learning. Then, we feed the augmented images through the model and calculate their Grad-CAM heatmaps to get $\{g_j(T_j(x_j)\}_{j=1}^n$. We assume one of the images $x_i$ where $i\in{1..n}$ is the query image and calculate its Grad-CAM heatmap without any transformation. We then apply the same transformation we had applied to $x_i$ to the Grad-CAM heatmap, instead of the image, to get $T_i(g_i(x_i))$.\par 
Our main idea is that if we transform an image, the interpretation should also be transformed in the same way. In addition, the interpretation should be specific to each image. We want $T_i(g_i(x_i))$ to be close to $g_i(T_i(x_i)))$ and far from $\{g_j(T_j(x_j)\}_{j \neq i}$. Hence, we define the following loss function:
\begin{equation}\nonumber
    L_i = -\text{log} \frac{\text{exp} \big(\text{sim}\big(T_i(g_i(x_i)), g_i(T_i(x_i))\big)/\tau \big)}{\sum_{j=1}^n \text{exp} \big(\text{sim}\big(T_i(g_i(x_i)), g_j(T_j(x_j)\big)/\tau\big)}
\end{equation} 

\noindent where $\tau$ is the temperature hyperparameter and $\text{sim}(a,b)$ measures a similarity between two heatmaps. In our experiments, we use cosine similarity. Note that cosine similarity is equivalent to L2 distance metric on normalized features. $g_i(.)$ always calculates the Grad-CAM of the top prediction of the original image regardless of the transformation since it is important to keep the category that the Grad-CAM heatmap is calculated on consistent for the positive pair involving the query image $x_i$.

We call our loss term Contrastive Grad-CAM Consistency Loss ($L_{CGC}$). This loss is similar to the standard contrastive self-supervised learning loss \cite{He2019MomentumCF} with two main differences: (1) Our loss is defined on the interpretation of the network output instead of the image features; (2) The interpretation of the original query image is also augmented with the same parameters to match the interpretation of the augmented image. This compensates for the transformation to make the interpretations aligned.

In practice, since we run the optimization on mini-batches, we assume each image is the query once and sum over all losses optimizing $L_{CGC} = \sum_i L_i$. This can be implemented efficiently for the whole mini-batch by augmenting each image once and calculating Grad-CAM for each image twice. Some contrastive self-supervised learning algorithms like MoCo \cite{He2019MomentumCF} use a memory bank to increase the number of negative pairs, but for simplicity, we do not use a memory bank and use mini-batches of size $256$. Thus, our method is more similar to SimCLR \cite{chen2020simple} than MoCo \cite{He2019MomentumCF}.

Our final loss is the combination of the standard cross-entropy loss ($L_{CE}$) and our contrastive Grad-CAM consistency loss ($L_{CGC}$). Hence, we minimize the following loss function:
\begin{equation}\nonumber
    L = L_{CE} + \lambda L_{CGC}
    \label{eq:our_final_loss}
\end{equation}
where, $\lambda$ is a hyper-parameter that controls the trade-off between the two loss terms. Note that our $L_{CGC}$ loss term does not use image labels as it uses pseudo labels for Grad-CAM. This enables us to use additional unlabeled data to improve both the accuracy and explainability of the resulting model in Section \ref{sec:semi_supervised_learning}.

\section{Experiments}
In this section, we perform a variety of experiments using our CGC method. For each of the experiments, `baseline' refers to a model trained from scratch using the standard cross-entropy loss unless noted otherwise.
We will report the classification accuracy along with the following metrics used for evaluating the explanation heatmaps: 

\noindent \textbf{Content Heatmap (CH):} Introduced in \cite{pillai2021explainable}, this metric is a measure of the summation of $\ell_1$-normalized heatmap contained within the annotated bounding box of the object. If the model interpretation is consistent with human annotations of the object location, we can assume that the percentage of the heatmap that lies inside the object annotation mask should be close to 100\%. Hence, we expect this metric to be high.

\noindent \textbf{CGC loss:} We also evaluate the explanation heatmaps using the same $L_{CGC}$ loss that we use for training our models. Although this loss is already used in our optimization, we believe it is important to show that the loss is small on the unseen test data as well, i.e., the method generalizes from the training to test set.
We use ImageNet \cite{ILSVRC15} validation set to report this metric. We use a batch size of 32 to compute this loss term. For every query image in the batch, we pair it with an augmentation of the corresponding query image as the positive pair and consider each of the remaining 31 images in the batch to be the negative pairs.

\noindent \textbf{Insertion AUC score (IAUC):} This metric \cite{Petsiuk2018rise} successively inserts pixels from the highest to lowest attribution scores and makes predictions. The area under the curve defined by the prediction scores is then defined as the IAUC score. We expect the IAUC score to be higher for a better interpretation. 
% Note that we do not report the deletion AUC (DAUC) since deleting the highest attribution pixels from an image would result in the model relying on new regions of the image for decision making.
\\

\noindent \textbf{Implementation Details:}
We use PyTorch \cite{Paszke2019PyTorchAI} to train and evaluate our models for all experiments. We use SGD (weight decay=$1e-4$, learning rate=$0.1$, momentum=$0.9$, and batch size=$256$) to optimize both ResNet18 and ResNet50 \cite{He2015DeepRL} models. We train ImageNet \cite{ILSVRC15} models for 90 epochs and decay the learning rate by 0.1 every $30$ epochs. For transformation $T_i$ in our loss term, we use standard data augmentations (scaling, flipping, and translation). The baseline models also use these same augmentations. We use ResNet50 architecture for all our experiments unless otherwise noted. Training our ResNet50 on 2 Titan-RTX GPUs takes approximately 100 hours, whereas training the baseline takes approximately 70 hours on the same setting. We use $\tau=0.5$ for all experiments using our method. For ImageNet dataset, we use $\lambda=1.0$ for our method using ResNet18 and $\lambda=0.5$ for ResNet50.

We first report our results on ImageNet and UnRel \cite{Peyre17} datasets. We then show results on fine-grained classification tasks including limited data settings. Finally, we discuss results using unlabeled data with our loss term.

\subsection{Contrastive Grad-CAM Consistency (CGC)}
\label{sec:gcam_cons_sup_learn}
We train a network from scratch using the CGC method (cross-entropy loss and contrastive loss) and compare it to a baseline network trained from scratch using cross-entropy loss only. We report the Top-1 classification accuracy as well as heatmap evaluation metrics. We show results for both the ImageNet and UnRel \cite{Peyre17} datasets in order to show that our approach generalizes across multiple datasets. For both datasets, on the ResNet50 architecture, we show that our model has a slight (less than 2\% points) decrease in classification accuracy but shows a significant (greater than 15\% points) improvement on the explanation metrics.

\noindent {\bf ImageNet:}
Quantitative results on ImageNet validation set are reported in Table \ref{tab:class_inet}. For the ResNet50 model, we have a marginal drop in classification accuracy (1.5 points), whereas CH increases by 17 points. Since, ImageNet is a large dataset, we do not expect our regularizer to improve the classification accuracy on ImageNet. Moreover, our main focus is on improving the consistency of the explanations and we are willing to accept a marginal drop in classification accuracy at the cost of improved explanation consistency.

Table \ref{tab:iauc_r50} reports the evaluation results using the IAUC metric \cite{Petsiuk2018rise} on the ImageNet validation set and shows that our method quantitatively improves the Grad-CAM explanation heatmaps of the underlying model.

\begin{table}[h!]
    \begin{center}
        \begin{tabular}{| l | c | c | c | c |}
        \hline
             Arch & Method & Top-1  & CH (\%) & CGC\\
             & & Acc (\%) & & Loss\\
             \hline
             \multirow{3}{*}{ResNet18} & Baseline & \textbf{69.76} & 54.47 & 3.19\\
             & GCC \cite{pillai2021explainable} & 67.74 & 57.73 & 3.14\\
             & Ours (CGC) & 66.37 & \textbf{65.83} & \textbf{2.59}\\
             \hline
             \multirow{3}{*}{ResNet50} & Baseline & \textbf{76.13} & 54.77 & 3.15\\
             & GCC \cite{pillai2021explainable} & 74.40 & 59.42 & 3.09\\
             & Ours (CGC) & 74.60 & \textbf{71.75} & \textbf{2.64}\\
        \hline
        \end{tabular}
    \end{center}
    \vspace{-0.15in}
    \caption{Classification accuracy along with the Content Heatmap (CH) and CGC Loss explanation metrics on ImageNet validation set. Note that lower is better for CGC Loss.}
    \label{tab:class_inet}
\end{table}

\begin{table}[h!]
    \begin{center}
        \begin{tabular}{|l|c|}
        \hline
            Method & Insertion AUC\\
            \hline
            Baseline & 0.4860\\
            Ours (CGC) & \textbf{0.5216}\\
            \hline
        \end{tabular}
    \end{center}
    \vspace{-0.15in}
    \caption{The Grad-CAM explanation maps generated by our ResNet50 model outperforms the baseline ResNet50 model on the IAUC metric \cite{Petsiuk2018rise} using the ImageNet validation set.}
    \label{tab:iauc_r50}
\end{table}

\noindent {\bf UnRel:}
The UnRel dataset \cite{Peyre17} consists of images capturing unusual relations between objects. These images were collected from the web using triplet queries such as `person ride giraffe'. This dataset thus captures objects occurring in unusual spatial configurations and contexts. Both the baseline and CGC models are trained on ImageNet and evaluated on the 28 object categories of the UnRel dataset that overlap with ImageNet. We report the corresponding quantitative results in Table \ref{tab:class_unrel}. We observe that although our CGC model was trained on ImageNet, which is an object centric dataset, the improvement in our explanation heatmap generalizes to objects occurring in unusual backgrounds and spatial configurations. 

\begin{table}[h!]
    \begin{center}
        \begin{tabular}{|l|c|c|}
        \hline
             Model & Top-1 Acc (\%) & CH (\%)\\\hline
             Baseline  & \textbf{40.66} & 51.66 \\
             Ours (CGC) & 38.25 & \textbf{74.20} \\
        \hline
        \end{tabular}
    \end{center}
    \vspace{-0.15in}
    \caption{Classification accuracy and Content Heatmap (CH) evaluation on the 28 UnRel categories \cite{Peyre17} that overlap with ImageNet.}
    \label{tab:class_unrel}
\end{table}

\noindent {\bf Generalization to another interpretation method:}
We also show that the improved explanations of our model trained using Grad-CAM generalize to Contrastive Top-down Attention (cMWP) \cite{Zhang2016TopDownNA}, which is another explanation method. We report CH evaluation results using cMWP in Table \ref{tab:ch_exc_backprop}.

\begin{table}[h!]
    \begin{center}
        \begin{tabular}{|l|c|}
        \hline
            Method & CH (\%) using cMWP \cite{Zhang2016TopDownNA}\\
            % & \cite{Zhang2016TopDownNA} \\
            \hline
            Baseline & 74.78\\
            GCC \cite{pillai2021explainable} & 75.08\\
            Ours (CGC) & \textbf{75.50}\\
            \hline
        \end{tabular}
    \end{center}
    \vspace{-0.15in}
    \caption{Our method generalizes to another explanation method, Contrastive Top-down Attention (cMWP) \cite{Zhang2016TopDownNA} although our model was trained using Grad-CAM. We report the CH metric computed using cMWP on ImageNet validation set. All models use ResNet50 architecture.}
    \label{tab:ch_exc_backprop}
\end{table}

\noindent \textbf{Role of negative heatmaps in $L_{CGC}$:}
As part of our $L_{CGC}$ loss, together with an augmented image as the positive pair, we use the rest of the images in the batch to compute the negative heatmaps. The presence of negative heatmaps is expected to encourage the model to learn explanation heatmaps specific to each image. This would result in the model learning features corresponding to the most discriminative regions of the object. If the negative heatmaps are not used as part of the contrastive loss, the model can cheat by learning a trivial solution for the explanation consistency and would result in the explanation heatmaps being spread uniformly across the image. We verify this be training a model where the $L_{CGC}$ uses a simple $\ell_2$ loss on the normalized heatmap of the positive pair and does not use negative heatmaps. We observe that the resulting model indeed results in explanation heatmaps which are diffused across the image as verified by the low CH in Table \ref{tab:class_ch_cgc_without_contrast}.

\begin{table}[H]
    \begin{center}
        % \begin{tabular}{|l|c|c|}
        \scalebox{0.8}{
        \begin{tabular}{|l|c|c|c|c|}
        \hline
        \multirow{2}{*}{Model} &  \multicolumn{2}{|  c  |}{ResNet18} & \multicolumn{2}{|  c  |}{ResNet50} \\ \cline{2-5}
             & Top-1 Acc (\%) & CH (\%) & Top-1 Acc (\%) & CH (\%)\\\hline
            %  Baseline  & \textbf{76.13} & 54.77 \\
             CGC & 66.37 & \textbf{65.83} & 74.60 & \textbf{71.75} \\
             CGC w/o neg & 67.00 & 44.08 & 74.80 & 39.84 \\
        \hline
        \end{tabular}
        }
    \end{center}
    \vspace{-0.15in}
    \caption{Comparison of ImageNet evaluation against model trained without using negative heatmaps as part of $L_{CGC}$.
    % Classification accuracy and Content Heatmap (CH) evaluation on ImageNet for $L_{CGC}$ without negative heatmaps. 
    The model trained without the negative heatmaps as part of $L_{CGC}$ loss results in a very low CH, thus confirming our hypothesis that the lack of negative heatmaps would result in a model learning a trivial solution of generating heatmaps diffused throughout the image.}
    \label{tab:class_ch_cgc_without_contrast}
\end{table}

\begin{table*}[ht!]
    \begin{center}
        \begin{tabular}{| l | c | c | c | c |}
        \hline
             Method & CUB-200 & FGVC-Aircraft & Cars-196 & VGG Flowers-102\\\hline
             Baseline & 80.09 $\pm$ 0.89 & 83.65 $\pm$ 0.15 & 89.71 $\pm$ 0.14 & \textbf{96.09 $\pm$ 0.23} \\
%             Stanford Dogs & 82.57 & 81.59 \\
             Ours (CGC) & \textbf{81.49 $\pm$ 0.09} & \textbf{85.72 $\pm$ 0.20} & \textbf{90.28 $\pm$ 0.08} & \textbf{96.18 $\pm$ 0.09}\\
        \hline
        \end{tabular}
    \end{center}
    \vspace{-0.15in}
    \caption{Evaluation of classification accuracy of the baseline against our CGC method for fine-grained datasets. We run 3 trials and report the mean and standard deviation. We observe consistent improvements in the classification accuracy across all four datasets with the largest gain on the FGVC-Aircraft dataset.}
    \vspace{-0.1in}
    \label{tab:class_fg}
\end{table*}

\begin{table*}[ht!]
    \begin{center}
       \begin{tabular}{|l|ccc|ccc|}
       \hline
         &  \multicolumn{3}{c|}{CUB200} & \multicolumn{3}{c|}{Cars196} \\
        Method & 1-shot & 5-shot & 10-shot & 1-shot & 5-shot & 10-shot \\
        \hline
        Baseline & 13.7 $\pm$ 0.3   & 51.7 $\pm$ 0.3 & 66.4 $\pm$ 0.2 & 6.1 $\pm$ 0.2 & 34.3 $\pm$ 0.4 & 61.1 $\pm$ 0.4 \\
        Ours (CGC) & \textbf{15.8 $\pm$ 0.3}  & \textbf{55.2 $\pm$ 0.3} & \textbf{68.4 $\pm$ 0.3} & \textbf{6.5 $\pm$ 0.2} & \textbf{36.9 $\pm$ 0.4} & \textbf{63.0 $\pm$ 0.4} \\
        \hline \hline
         &  \multicolumn{3}{c|}{Aircrafts} & \multicolumn{3}{c|}{Flowers} \\
         &  1-shot & 5-shot & 10-shot & 1-shot & 5-shot & 10-shot \\
        \hline
        Baseline & 7.7 $\pm$ 0.3 & 25.7 $\pm$ 0.4 & 41.4 $\pm$ 0.3 & 52.1 $\pm$ 0.5 & \textbf{85.6 $\pm$ 0.4} & \textbf{93.2 $\pm$ 0.2} \\
        Ours (CGC) & \textbf{8.0 $\pm$ 0.3} & \textbf{26.9 $\pm$ 0.4} & \textbf{42.9 $\pm$ 0.3} & \textbf{53.3 $\pm$ 0.5} & \textbf{85.8 $\pm$ 0.4} & \textbf{93.4 $\pm$ 0.2}  \\
        \hline
        \end{tabular}
    \end{center}
    \vspace{-0.15in}
    \caption{Classification accuracy for the limited-data setting on the fine-grained datasets. We run 20 trials and report the mean and standard deviation. We observe consistent improvement across all datasets with the largest improvements observed on the CUB-200 and Cars-196 datasets.}
    \vspace{-0.1in}
    \label{tab:finegrained_limited}
\end{table*}

\begin{table*}[ht!]
    \begin{center}
       \begin{tabular}{|l|cccc|cccc|}
       \hline
         &  \multicolumn{4}{c|}{CUB-200} & \multicolumn{4}{c|}{Cars-196} \\
        Method & 1-shot & 5-shot & 10-shot & Full training set & 1-shot & 5-shot & 10-shot & Full training set\\
        \hline
        Baseline & 55.54 & 57.60 & 61.39 & 63.71 & \textbf{60.51} & \textbf{64.08} & 64.31 & 65.58 \\
        Ours (CGC) &  \textbf{61.63} & \textbf{63.55} & \textbf{73.86} & \textbf{71.08} & 60.48 & 62.13 & \textbf{64.63} & \textbf{69.04}\\
        \hline
        \end{tabular}
    \end{center}
    \vspace{-0.15in}
    \caption{Content Heatmap (CH) evaluation results for limited and full training data settings on CUB-200 and Cars-196 datasets. CH results are better for all CUB-200 sets but are marginally worse for 1-shot and 5-shot settings on Cars-196.  We believe the lower accuracy of the model for few-shot setting results in noisy interpretations, so the CH metric becomes less reliable.}
    \vspace{-0.15in}
    \label{tab:content_heatmap_fgfs}
\end{table*}

\subsection{Fine-Grained Classification}
\label{sec:fine_grained_classification}
We now report the results for fine-grained classification on CUB-200 \cite{welinder2010caltech}, FGVC-Aircraft \cite{maji13fine-grained}, Stanford Cars-196 \cite{KrauseStarkDengFei-Fei_3DRR2013}, and VGG Flowers \cite{Nilsback08} datasets. All models used for fine-grained classification are pretrained on ImageNet and all layers are fine-tuned for the new dataset. \par 

We find that our method can be used as a form of regularization, which is particularly helpful in these fine-grained classification scenarios. While our method trains to produce better interpretation, using a contrastive loss on the interpretation will allow the model to learn unique attention for individual classes. Note that the negative samples in the contrastive loss encourages the interpretation to be different from other samples. As a result, the trained model limits the attention to the most discriminative part(s) of the object and hence, our method acts like a regularizer when adopted for fine-grained classification tasks.

For our CGC method, we use $\lambda=0.25$ for the CUB-200 and Cars-196 datasets, and $\lambda=0.5$ for the FGVC-Aircraft and VGG Flowers datasets. 
 
Results for classification accuracy are shown in Table  \ref{tab:class_fg}, and results for Content Heatmap evaluation are shown in Table \ref{tab:content_heatmap_fgfs}. For all fine-grained datasets, the model trained with our CGC method achieves both improved classification accuracy and improved explanation scores. For most datasets, improvements to classification accuracy is marginal, but the FGVC-Aircraft dataset shows an improvement of over 2\% points. This experiment demonstrates that our method not only improves explainability, but also classification accuracy for datasets that require the network to focus on the ``most discriminative'' features of an object.

\noindent {\bf Limited Training Data:}
Using our method for regularization can be particularly helpful when the amount of training data is limited. We evaluate our method with the same fine-grained classification settings, but limit the amount of training data for each class. We evaluate 1, 5, and 10-shot training settings. We repeat our experiment for 20 episodes and report mean and standard deviation over all episodes. In each episode, we randomly select $n \in \{1,5,10\}$ samples from each class as training set and use the rest of the samples as a validation set. We initialize the model from a ResNet50 model pretrained on ImageNet with categorical cross-entropy loss only, then finetune all layers on the limited training set. We use the same random seed for our method as well as the baseline. 
For our method in this limited-data setting, we use $\lambda=0.8$ for the CUB-200, Cars-196, and VGG Flowers datasets and $\lambda=0.6$ for the FGVC-Aircraft datasets. For all datasets we use $\tau=0.5$ as the value for the temperature hyperparameter.

Results for classification accuracy in this limited-data setting are found in Table \ref{tab:finegrained_limited}. Both baseline and CGC models perform better with more samples per class, but our method consistently outperforms the baseline. The magnitude of improvement varies from ~1-4\% points. Notably, our method outperforms the baseline by 4\% points (5-shot and 10-shot) on CUB-200 and by 2\% points (5-shot and 10-shot) on Cars-196. Since only CUB-200 and Cars-196 contain bounding box annotations, we report the CH evaluation metric for these two datasets in Table \ref{tab:content_heatmap_fgfs}. CH results are better for all CUB-200 sets but are marginally worse for 1-shot and 5-shot settings on Cars-196.  We believe the lower accuracy of the model for the few-shot setting results in noisy interpretations, so the CH metric becomes less reliable.

\begin{figure*}[ht!]
\centering
\begin{tabular}{|c|c|c|c|c|c||c|}
\hline
 \small{Polaroid} & \small{Boathouse} & \small{Great White} & \small{Cowboy} & \small{Snowmobile} & \small{Volleyball} & \small{Dogsled}\\
 \small{Camera} & & \small{Shark} & \small{Hat} & & & \\
%  \small{Tennis} & \small{Boathouse} & \small{Great White} & \small{Panda} & \small{Banana} & \small{Volleyball} & \small{Dogsled}\\
%  \small{Ball} & & \small{Shark} & & & & \\
\hline
    \begin{sideways} \quad \quad \scriptsize{Original} \end{sideways}    
    \includegraphics[width=.11\textwidth]{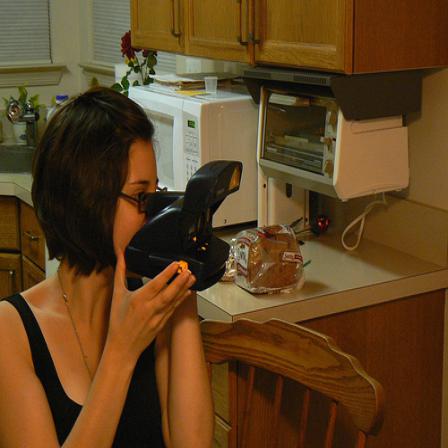}&
    \includegraphics[width=.11\textwidth]{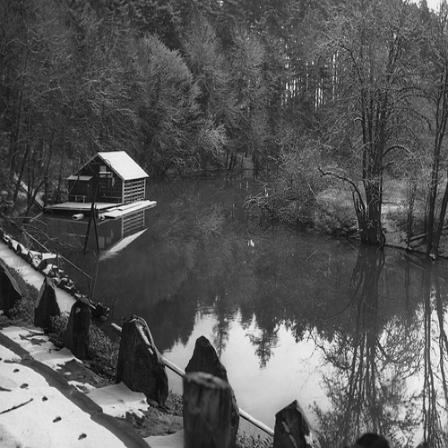}&
    \includegraphics[width=.11\textwidth]{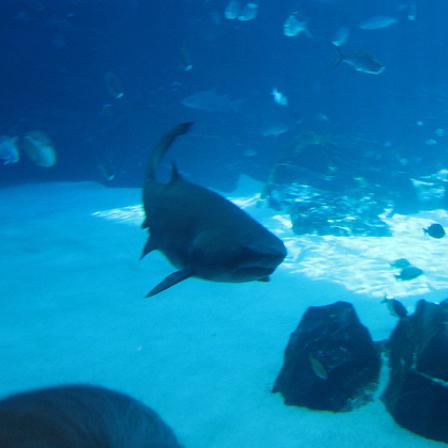}&
    \includegraphics[width=.11\textwidth]{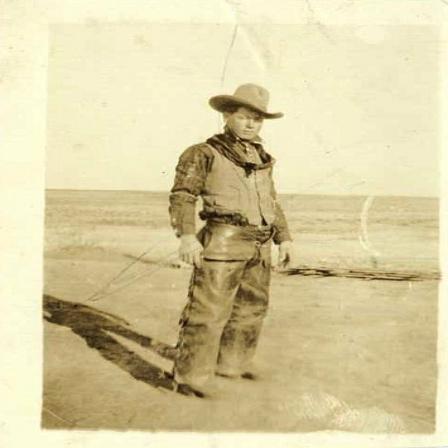}&
    \includegraphics[width=.11\textwidth]{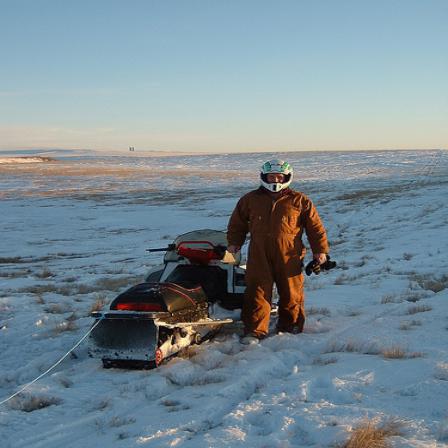}&
    \includegraphics[width=.11\textwidth]{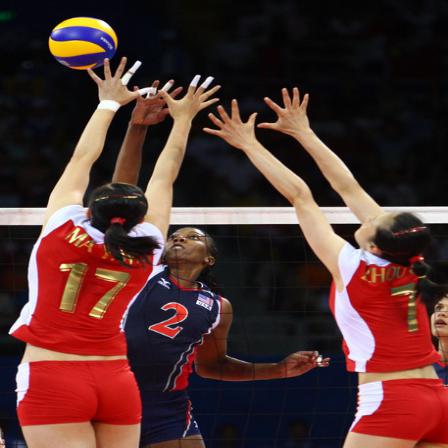}&
    \includegraphics[width=.11\textwidth]{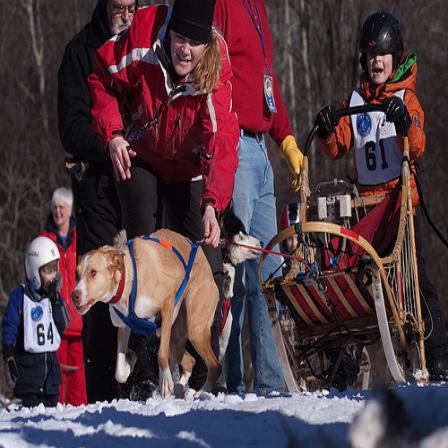}\\

    \begin{sideways} \hspace{0.25em} \scriptsize{Baseline G-CAM} \end{sideways}    
    \includegraphics[width=.11\textwidth]{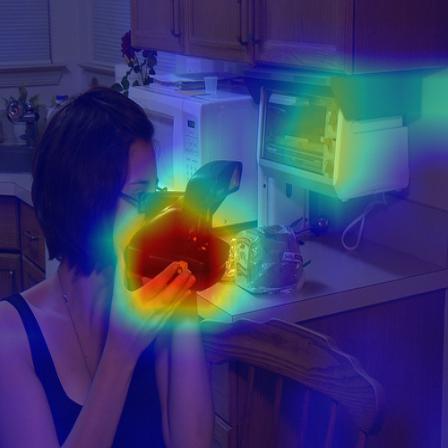}&
    \includegraphics[width=.11\textwidth]{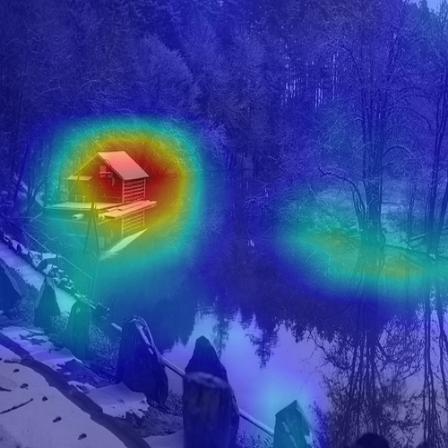}&
    \includegraphics[width=.11\textwidth]{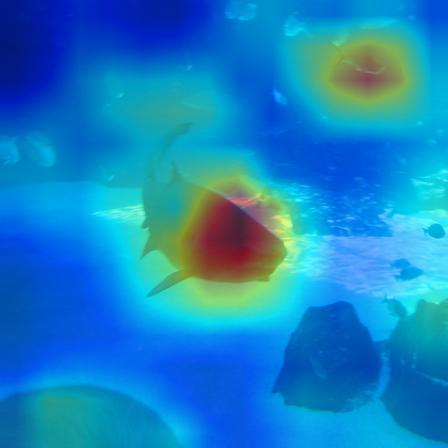}&
    \includegraphics[width=.11\textwidth]{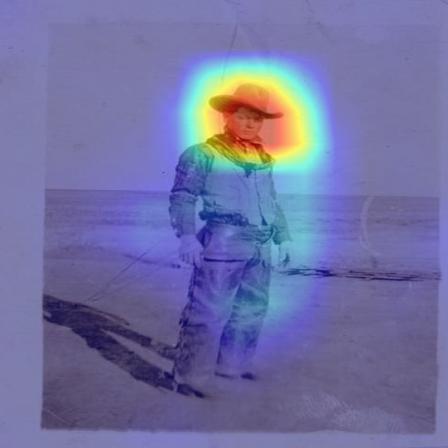}&
    \includegraphics[width=.11\textwidth]{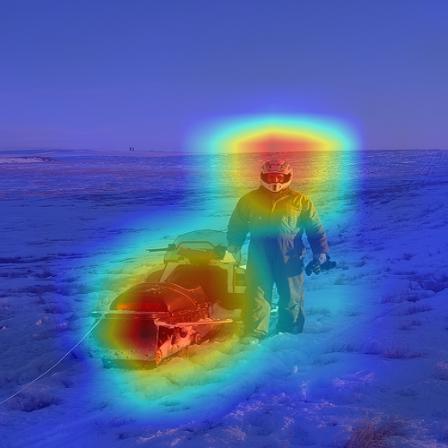}&
    \includegraphics[width=.11\textwidth]{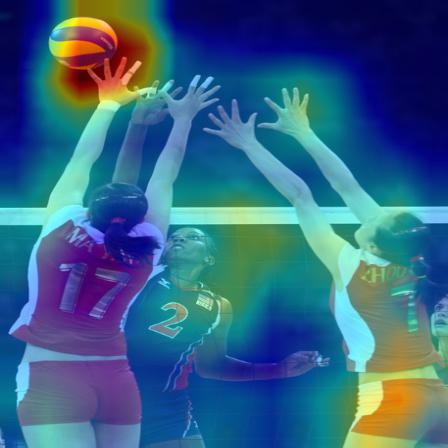}&
    \includegraphics[width=.11\textwidth]{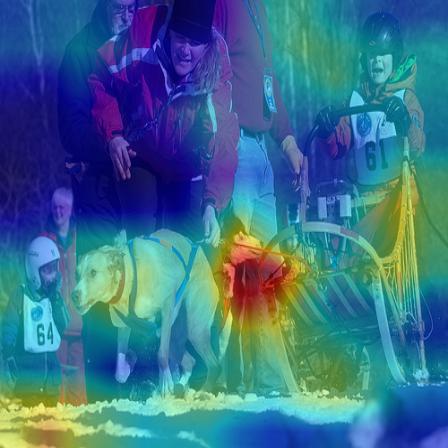}\\ 

    \begin{sideways} \quad \scriptsize{Ours G-CAM} \end{sideways}    
    \includegraphics[width=.11\textwidth]{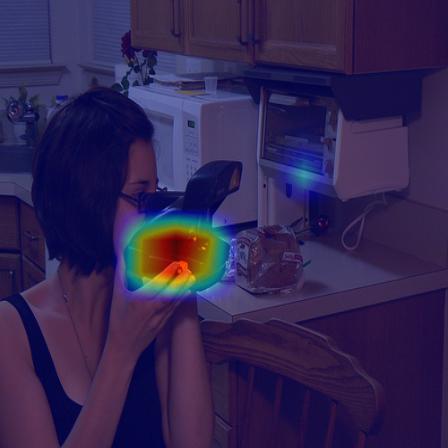}&
    \includegraphics[width=.11\textwidth]{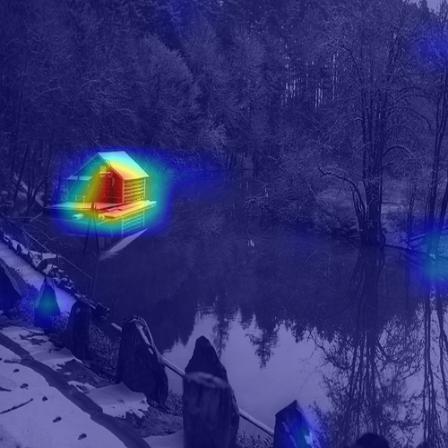}&
    \includegraphics[width=.11\textwidth]{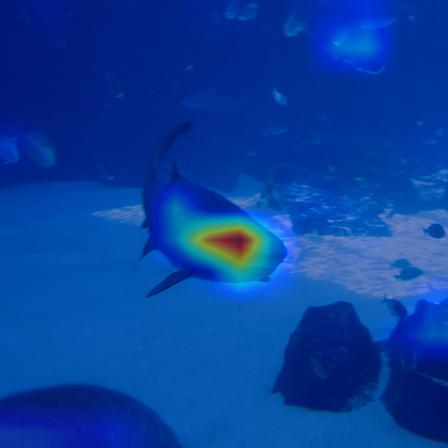}&
    \includegraphics[width=.11\textwidth]{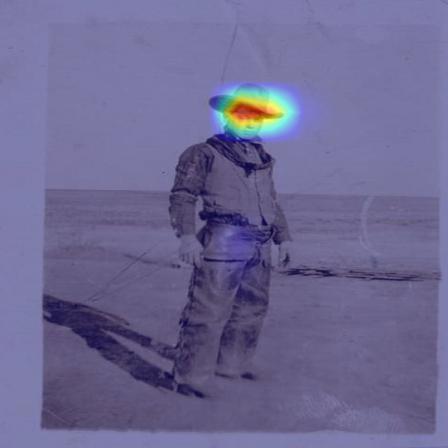}&
    \includegraphics[width=.11\textwidth]{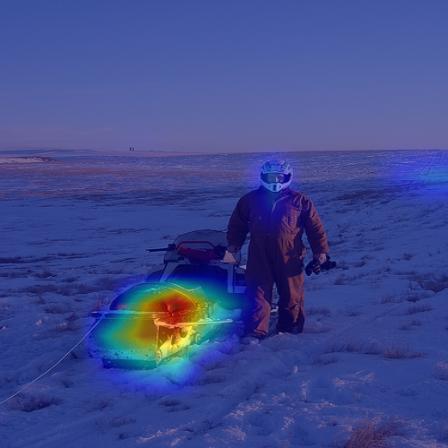}&
    \includegraphics[width=.11\textwidth]{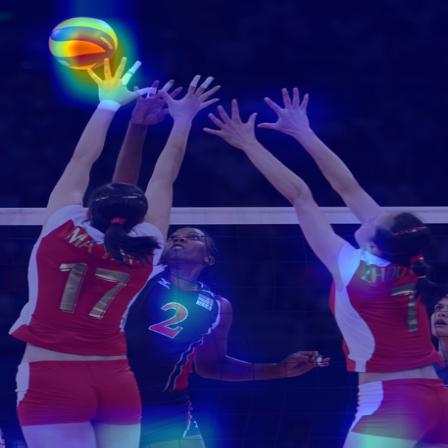}&
    \includegraphics[width=.11\textwidth]{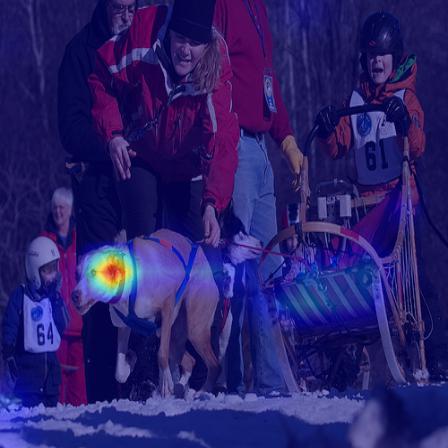}\\

\hline
\end{tabular}
\vspace{-0.05in}
\caption{Grad-CAM visualization results for images from ImageNet validation set using ResNet50. We observe that the model trained using our method focuses on the most discriminative regions of the object instead of the background pixels. In the 6th column above, we compute the Grad-CAM explanation for the category ``volleyball'' and we see that our model no longer focuses on the players and correctly highlights the volleyball as the cause of the target category. The last column contains a failure example for the category ``Dogsled''. The baseline model correctly highlights the sled whereas our model incorrectly highlights the dog.}
\label{fig:imagenet_gradcam_r50_vis1}
\end{figure*}

\subsection{Using Unlabeled Data for CGC Loss}
\label{sec:semi_supervised_learning}
Models trained with categorical cross-entropy loss only have no way to incorporate information from unlabeled data. 
Since our method does not rely on labels for computing the explanation heatmaps, we extend our method to use mostly unlabeled data with just a small fraction of labeled data for training. We use $1\%$ of ImageNet training images as labeled data and leverage the rest of ImageNet training images as the unlabeled data for the CGC loss term.
We initialize both the baseline and our method from a ResNet50 model trained using SwAV \cite{caron2020unsupervised}. 

% Results for classification accuracy after using this 1\% susbet of ImageNet training data are shown in Table \ref{tab:class_unlab}. Our method outperforms the baseline by 1\% point on both Top-1 and Top-5 accuracy metrics. 
Table \ref{tab:class_unlab} shows that our method outperforms the baseline by 1\% point on both the Top-1 and Top-5 accuracy metrics. Since we are using only 1\% labeled data, the resulting Grad-CAM heatmap will be less accurate and hence the 1\% point improvement is not insignificant. Even though our model is only able to leverage the additional unlabeled data for the CGC loss term, and not the categorical accuracy, we can see that the CGC loss acts as a regularizer to improve the generalization of the model.

\begin{table}[h!]
    \begin{center}
        \begin{tabular}{|l|c|c|c|}
        \hline
             Model & Top-1 (\%) & Top-5 (\%) & CH (\%)\\\hline
              Baseline & 54.00 & 78.69 & 46.08\\
              Ours (CGC) & \textbf{55.18} & \textbf{79.12} & \textbf{46.76}\\
             
        \hline
        \end{tabular}
    \end{center}
    \vspace{-0.15in}
    \caption{On the 1\% ImageNet limited-label setting, our CGC method leverages unlabeled data for the $L_{CGC}$ loss term and is able to improve the classification accuracy and explanations when evaluated on ImageNet validation data. Both models are initialized from a ResNet50 model trained in an unsupervised manner using SwaV \cite{caron2020unsupervised}.}
    \label{tab:class_unlab}
\vspace{-0.15in}
\end{table}

% \subsection{Background Challenges Evaluation}

\subsection{Ablation on $\lambda$}
We perform an ablation experiment to study the sensitivity of our method to the $\lambda$ hyperparameter. If $\lambda$ is too low, the cross-entropy term will dominate the optimization and thus the resulting improvement in the explanation consistency will be marginal, whereas if $\lambda$ is too high, the CGC loss will be applied on noisy heatmaps resulting in lower accuracy and lower explanation consistency. Table \ref{tab:ablation_lambda} shows the Top-1 accuracy and CH scores for different values of $\lambda$ for ResNet18 on ImageNet dataset. We choose $\lambda=1.0$ for ResNet18 as the best trade-off between accuracy and CH.
\begin{table}[!h]
    \centering
    \begin{tabular}{c | c | c | c | c | c | c}
         \small$\lambda$ & \small0 & \small0.01 & \small0.1 & \small0.5 & \small1.0 & \small2.0\\
         \hline
         \small{Top-1 Acc} & \small69.76 & \small66.97 & \small66.90  & \small65.14 & \small66.37 & \small65.89\\
         \hline
         CH & \small54.47 & \small56.18 & \small61.97 & \small66.19 & \small65.83 & \small52.60\\
    \end{tabular}
    \caption{Ablation to study the sensitivity of our method to the $\lambda$ hyperparameter for ResNet18 on ImageNet dataset.}
    \vspace{-0.15in}
    \label{tab:ablation_lambda}
\end{table}

\subsection{Qualitative Results}
Figure \ref{fig:imagenet_gradcam_r50_vis1} compares the Grad-CAM heatmaps generated by the baseline model against our CGC model on ImageNet. 
Our model consistently focuses on the discriminatory parts of the object and does not highlight background pixels. Additional results for CUB-200, Cars-196 and Aircrafts datasets are included in the appendix.

\section{Conclusion}
We introduce a contrastive learning method for improving the explanations generated by a deep neural network, training them to be consistent with spatial transformations. We emphasize the importance of evaluating the network based on its quality of explanation, and not only classification accuracy. Our CGC method significantly improves the explanation heatmaps while obtaining comparable classification accuracy on ImageNet and UnRel datasets. Furthermore, our method is able to boost the classification accuracy on fine-grained classification datasets such as CUB-200, Cars-196, VGG Flowers-102, and FGVC-Aircraft while improving the consistency of explanation heatmaps with human annotations. This demonstrates that our method acts as a regularizer that focuses more attention on the discriminating aspects of the image. We also show that our method is able to leverage unlabeled data to improve the classification accuracy in limited-label data settings.

\noindent \textbf{Limitations:}
Our method uses Grad-CAM \cite{Selvaraju2016GradCAMVE} algorithm to compute the explanation heatmaps for the original set of images as well as the augmented images, which are then used to compute the contrastive loss term $L_{CGC}$. In comparison to standard training with cross-entropy loss, our method requires additional compute to account for storing the additional gradient graph in memory during backpropagation. While this is an overhead during the training stage, we believe the incurred compute cost is offset by the improved explainability of the resulting model.

\noindent \textbf{Ethics Statement:}
Our method improves the explainability of image classification models and thereby increases trust and transparency of the underlying decision making process. However, our method is a data-driven approach and hence could reflect potential negative biases present in the training dataset. Moreover, explanation methods such as Grad-CAM \cite{Selvaraju2016GradCAMVE} can be an unreliable estimate of model interpretability (i.e., evidence for an incorrect prediction looking identical to evidence for a correct prediction). 

\noindent {\bf Acknowledgment:} 
This material is based upon work partially supported by the U.S. Air Force under Contract No. FA8750‐19‐C‐0098, U.S. Department of Commerce, National Institute of Standards and Technology under award number 60NANB18D279, NSF grant numbers 1845216 and 1920079, and funding from Northrop Grumman and SAP SE. Any opinions, findings, and conclusions or recommendations expressed in this material are those of the authors and do not necessarily reflect the views of the U.S. Air Force, DARPA, or other funding agencies. We would also like to thank the reviewers for their valuable feedback.

{\small
\bibliographystyle{ieee_fullname}
\bibliography{cvpr22}
}

\clearpage
\onecolumn
\section{Appendix}

\begin{figure}[h!]
\setlength{\linewidth}{\textwidth}
\setlength{\hsize}{\textwidth}
\centering
\begin{tabular}{|c|c|c|c|c|c|}
\hline
 \scriptsize{Seaside Sparrow} & \scriptsize{Anna Hummingbird}  & \scriptsize{Eared Grebe} &  \scriptsize{Pileated} & \scriptsize{Red Cockaded}   & \scriptsize{Red headed}  \\
  &  &  & \scriptsize{Woodpecker} & \scriptsize{Woodpecker} & \scriptsize{Woodpecker}\\
\hline
    \begin{sideways} \quad \hspace{0.25em}  \scriptsize{Original} \end{sideways}
    \includegraphics[width=.15\textwidth]{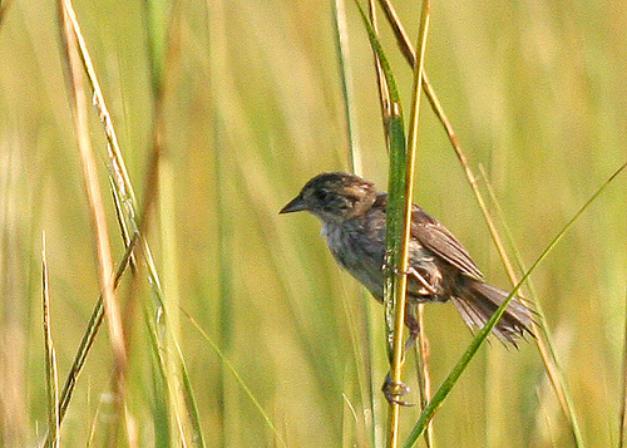}&
    \includegraphics[width=.15\textwidth]{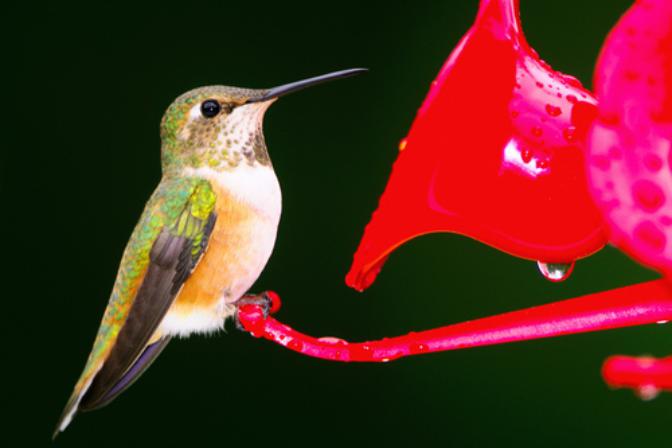}&
    \includegraphics[width=.15\textwidth]{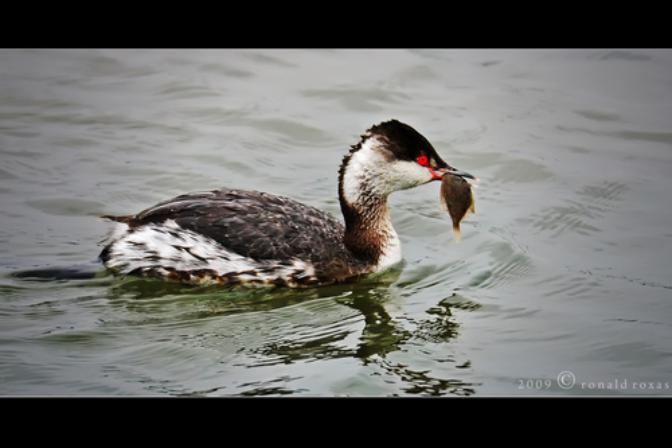}&
    \includegraphics[height=.12\textwidth]{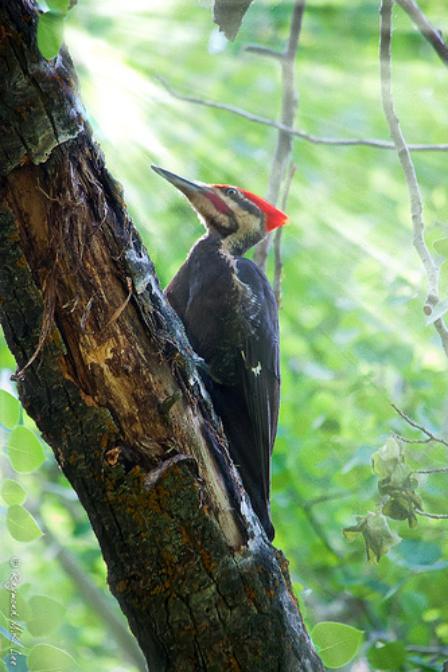}&
    \includegraphics[width=.15\textwidth]{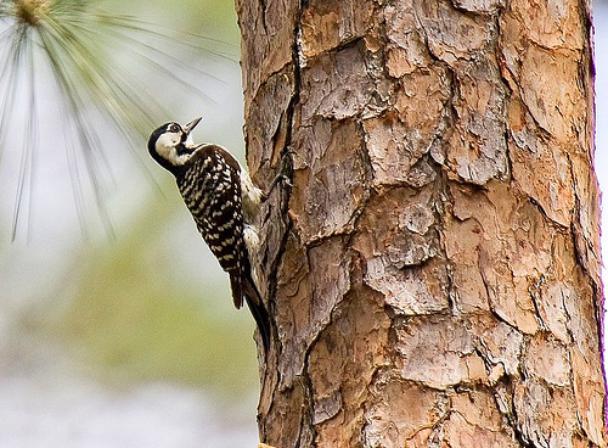}&
    \includegraphics[height=.12\textwidth]{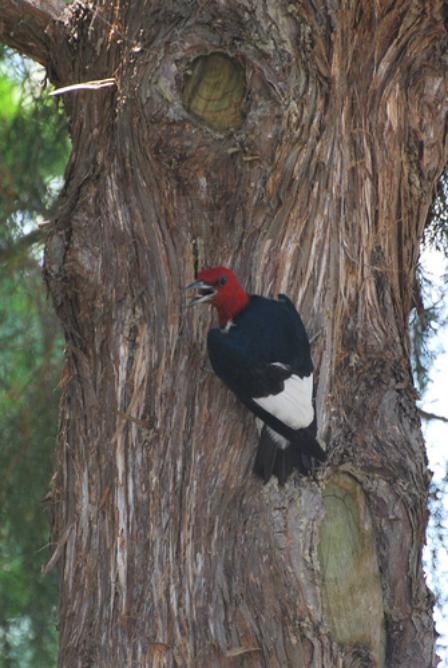}\\

    \begin{sideways} \hspace{0.1em} \scriptsize{Baseline G-CAM} \end{sideways}    
    \includegraphics[width=.15\textwidth]{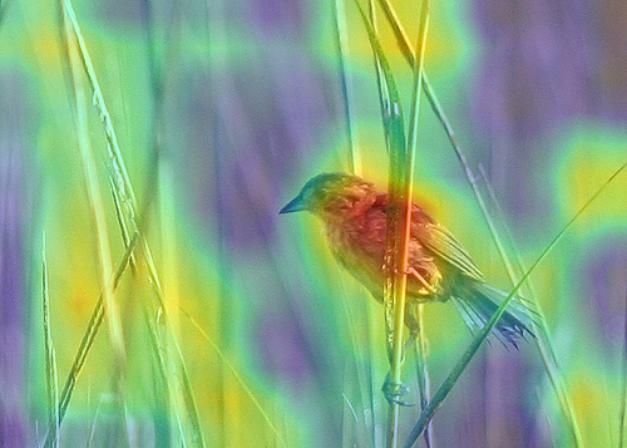}&
    \includegraphics[width=.15\textwidth]{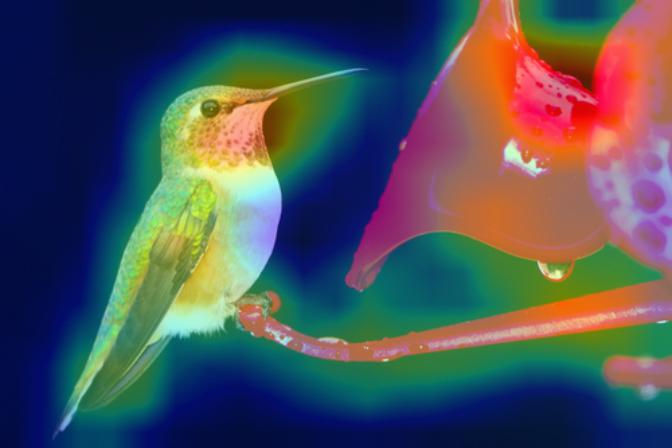}&
    \includegraphics[width=.15\textwidth]{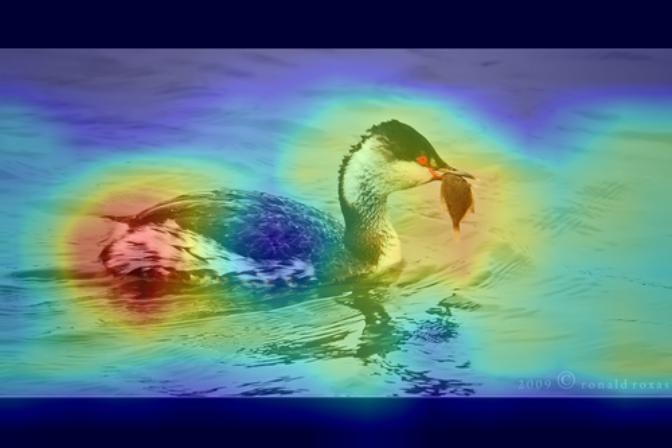}&
    \includegraphics[height=.12\textwidth]{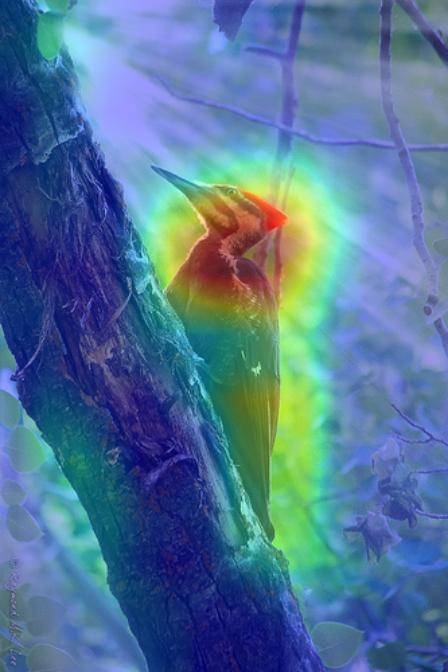}&
    \includegraphics[width=.15\textwidth]{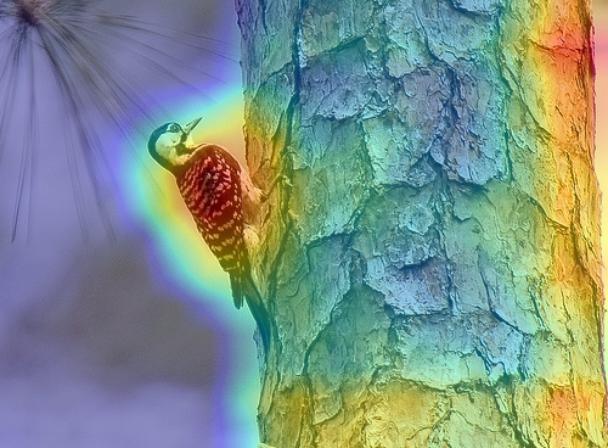}&
    \includegraphics[height=.12\textwidth]{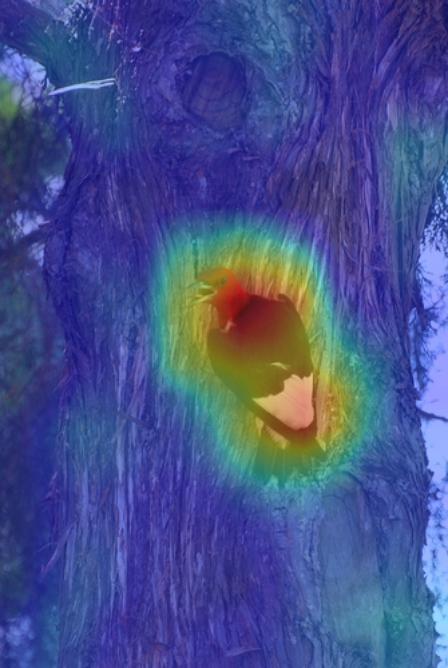}\\
    
    \begin{sideways} \hspace{0.25em} \scriptsize{Ours G-CAM} \end{sideways}    
    \includegraphics[width=.15\textwidth]{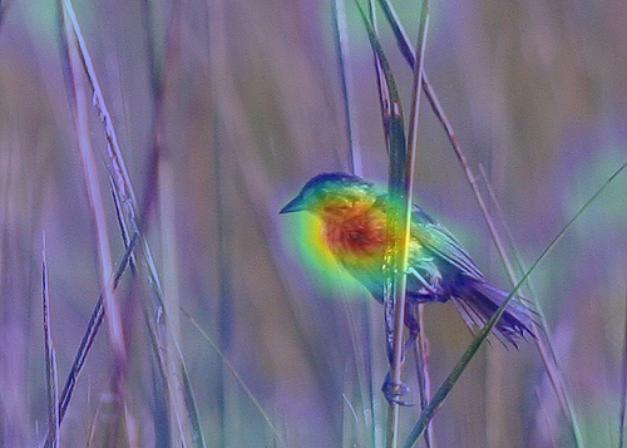}&
    \includegraphics[width=.15\textwidth]{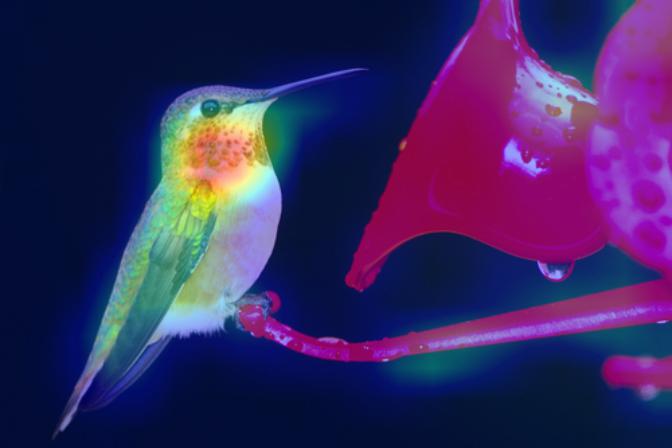}&
    \includegraphics[width=.15\textwidth]{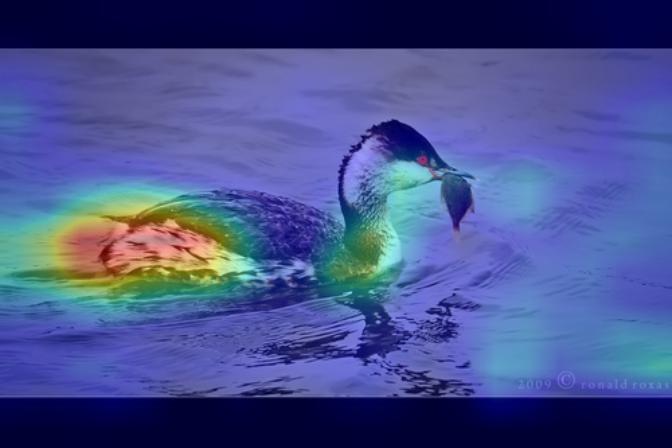}&
    \includegraphics[height=.12\textwidth]{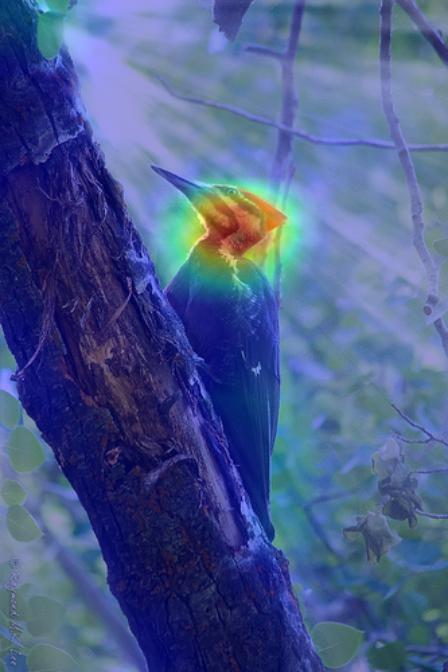}&
    \includegraphics[width=.15\textwidth]{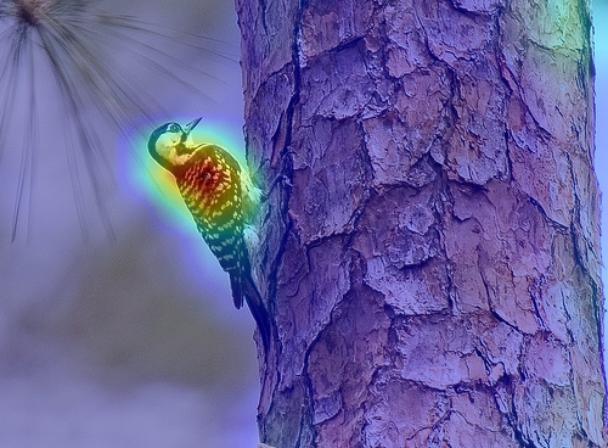}&
    \includegraphics[height=.12\textwidth]{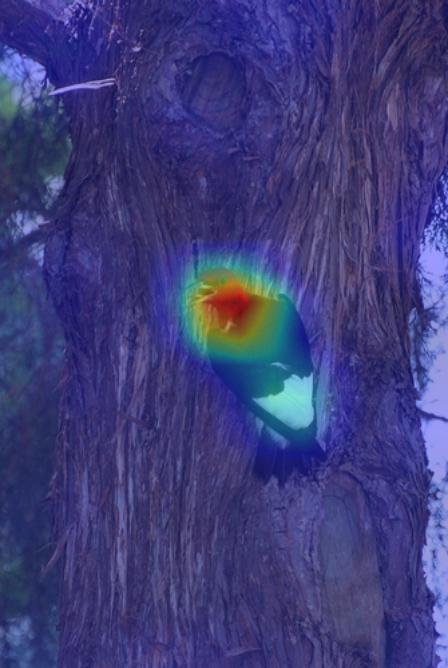}\\

\hline
\end{tabular}
\caption{Grad-CAM visualization results for images from the CUB-200 validation set using ResNet50. Interestingly, on the right column, the baseline is focusing on the whole bird while our method focuses on the head of the bird only. Given the name of the category ``Red headed Woodpecker'', it makes sense that the head should be the most discriminative region.}
\label{fig:birds_gradcam_r50_vis1}
\end{figure}

\begin{figure}[h!]
\setlength{\linewidth}{\textwidth}
\setlength{\hsize}{\textwidth}
\centering
\scriptsize
\begin{tabular}{|c|c|c|c||c|}
\hline
 \scriptsize{BMW ActiveHybrid 5} & \scriptsize{Lamborghini}  & \scriptsize{Porsche}  & \scriptsize{Jeep} & \scriptsize{Mercedes-Benz}\\
 \scriptsize{Sedan} & \scriptsize{Reventon Coupe} & \scriptsize{Panamera Sedan} & \scriptsize{Patriot SUV} & \scriptsize{E-Class Sedan 2012}\\
\hline
    \begin{sideways} \quad \quad \scriptsize{Original} \end{sideways}    
    \includegraphics[width=.17\textwidth]{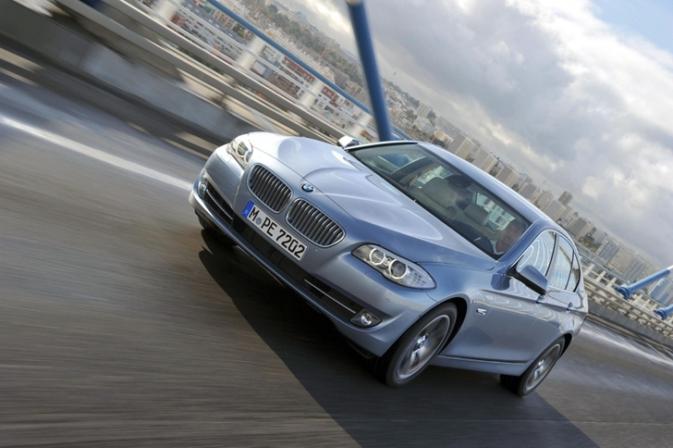}&
    \includegraphics[width=.2\textwidth]{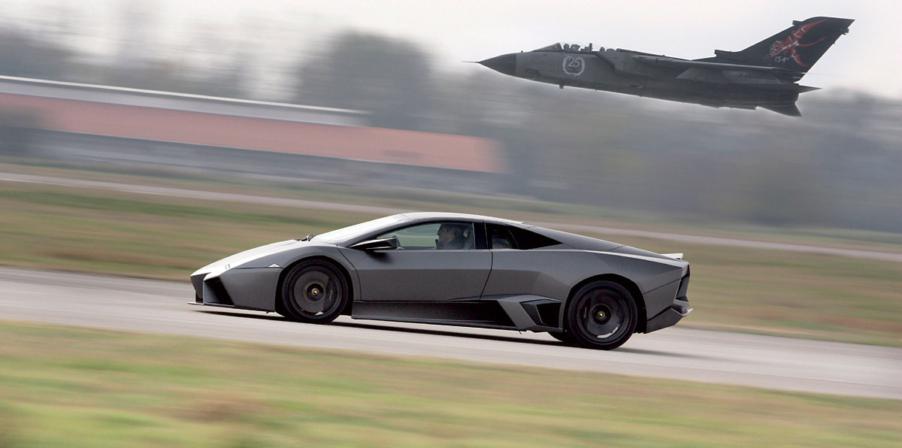}&
    \includegraphics[width=.16\textwidth]{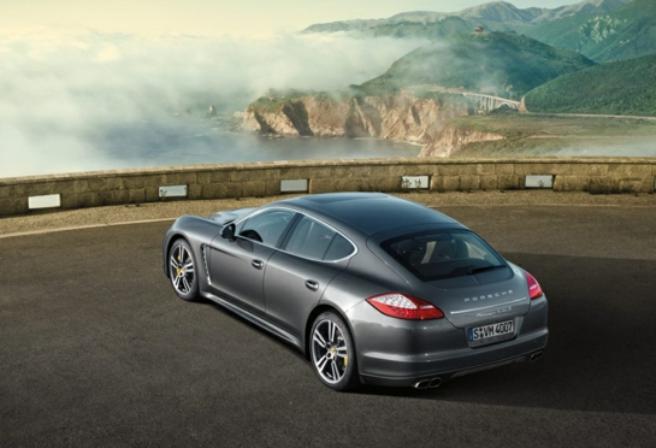}&
    \includegraphics[width=.16\textwidth]{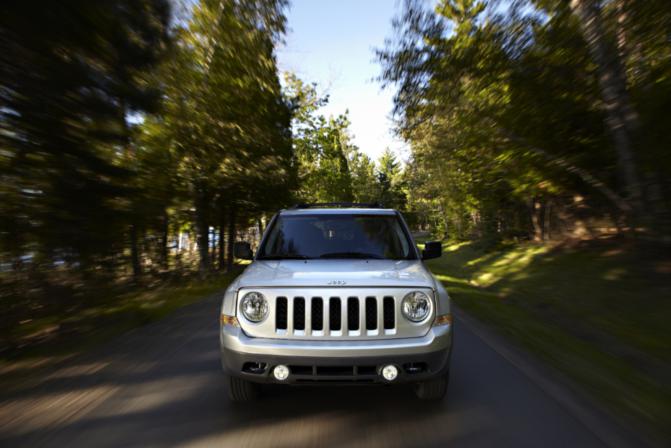}&
    \includegraphics[width=.16\textwidth]{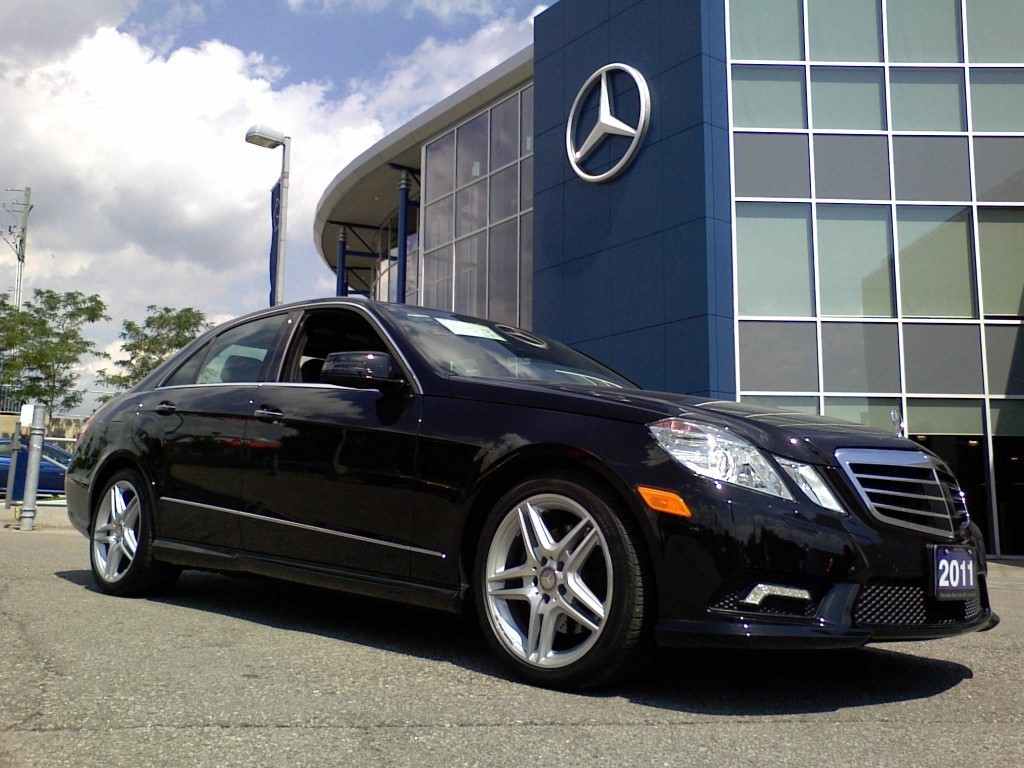}\\

    \begin{sideways} \hspace{0.25em} \scriptsize{Baseline G-CAM} \end{sideways}    
    \includegraphics[width=.17\textwidth]{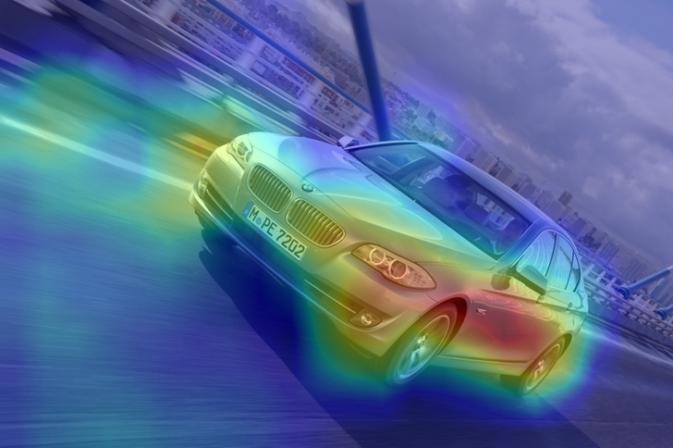}&
    \includegraphics[width=.2\textwidth]{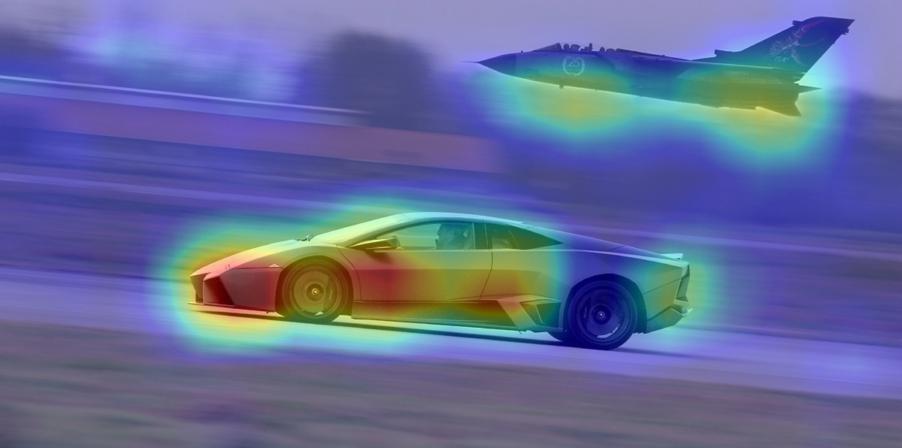}&
    \includegraphics[width=.16\textwidth]{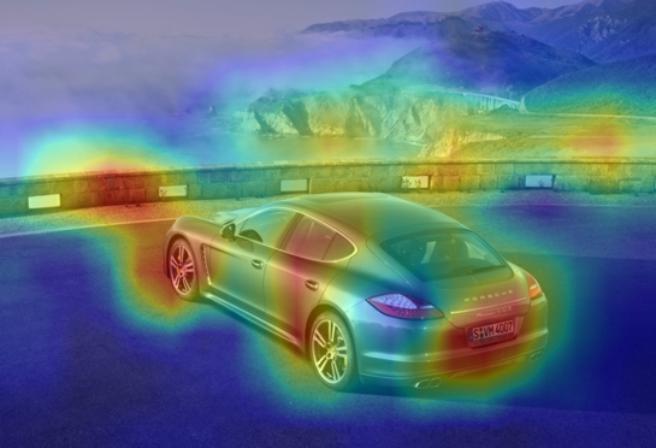}&
    \includegraphics[width=.16\textwidth]{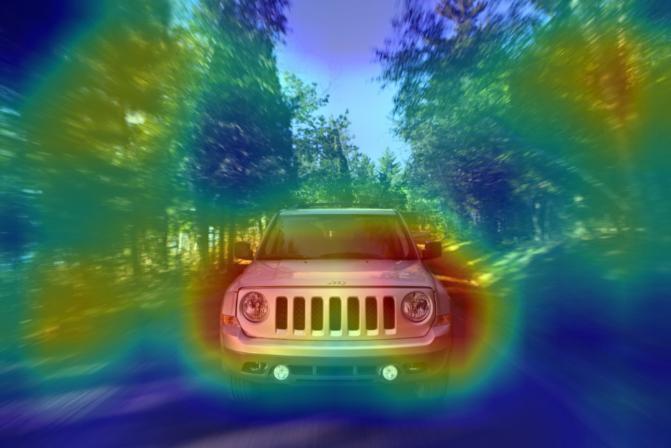}&
    \includegraphics[width=.16\textwidth]{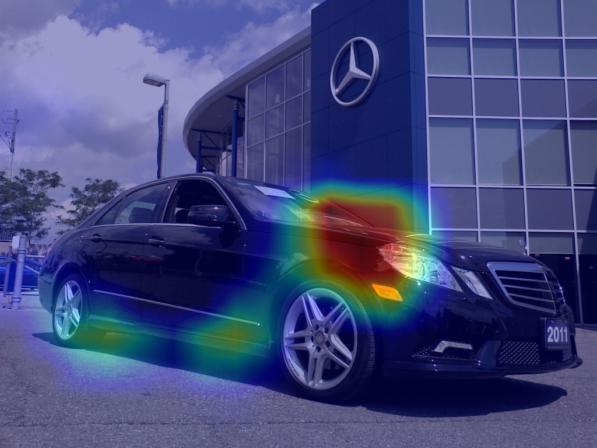}\\
    
    \begin{sideways} \quad \scriptsize{Ours G-CAM} \end{sideways}    
    \includegraphics[width=.17\textwidth]{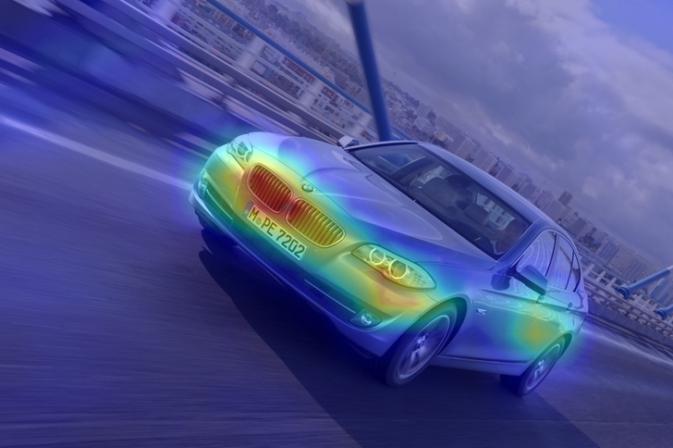}&
    \includegraphics[width=.2\textwidth]{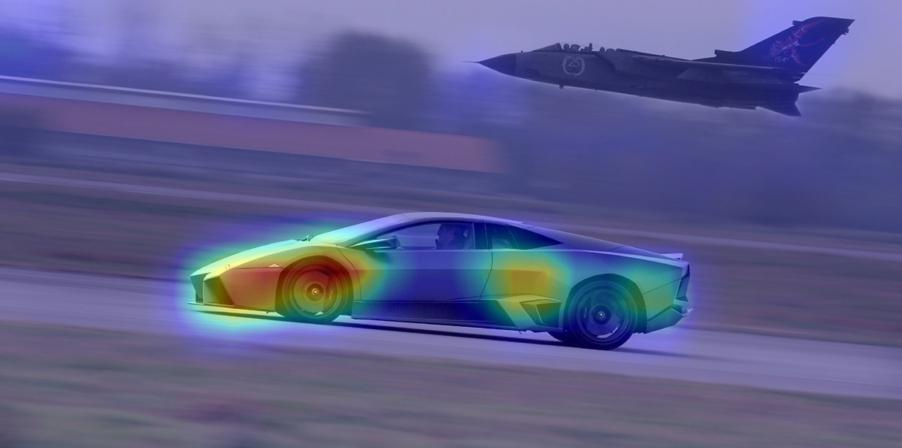}&
    \includegraphics[width=.16\textwidth]{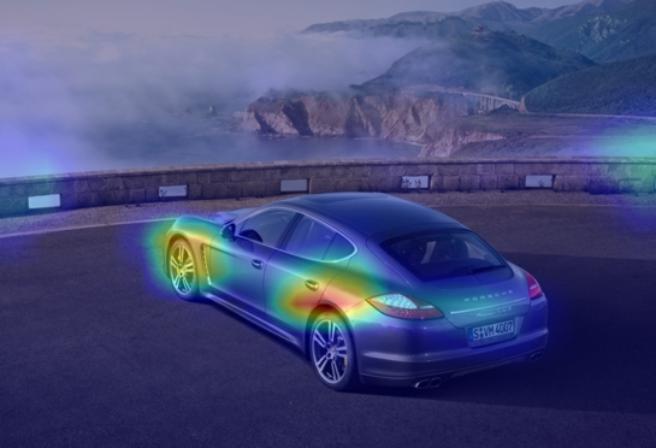}&
    \includegraphics[width=.16\textwidth]{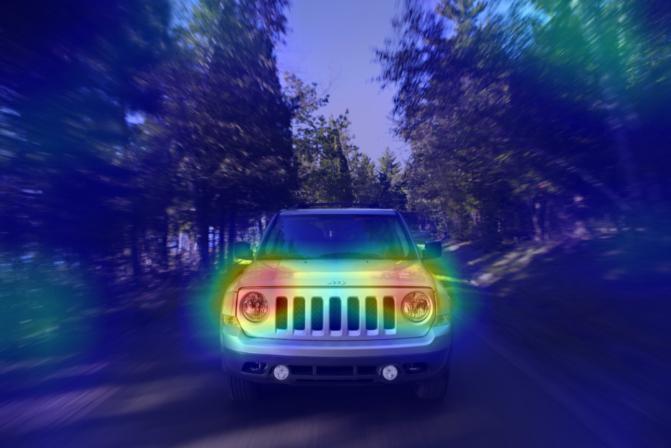}&
    \includegraphics[width=.16\textwidth]{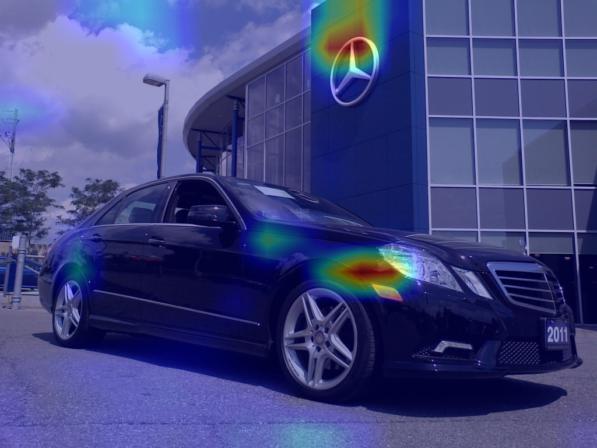}\\
\hline
\end{tabular}
\caption{Grad-CAM visualization results for images from the Cars-196 validation set using ResNet50. In the second column above, we see that our model is able to correctly focus on the object regions of ``Lamborghini'' whereas the baseline incorrectly highlights the fighter jet as well. The last column shows a failure example where our model incorrectly focuses on the ``Mercedes'' logo on the building instead of the car itself. This is an interesting failure case since the logo is a valid discriminatory attribute for a fine-grained car classification.}
\label{fig:cars_gradcam_r50_vis1}
\end{figure}

\clearpage
\begin{figure}
\setlength{\linewidth}{\textwidth}
\setlength{\hsize}{\textwidth}
\centering
\scriptsize
\begin{tabular}{|c|c|c|c|c||c|}
\hline
 \small{DHC-6} & \small{DC-8} & \small{DHC-6} & \small{A380} & \small{A340-500} & \small{Hawk T1} \\
 \small{} & & \small{} & & & \\
\hline
    \begin{sideways} \quad \scriptsize{Original} \end{sideways}    
    
    \includegraphics[width=.14\textwidth]{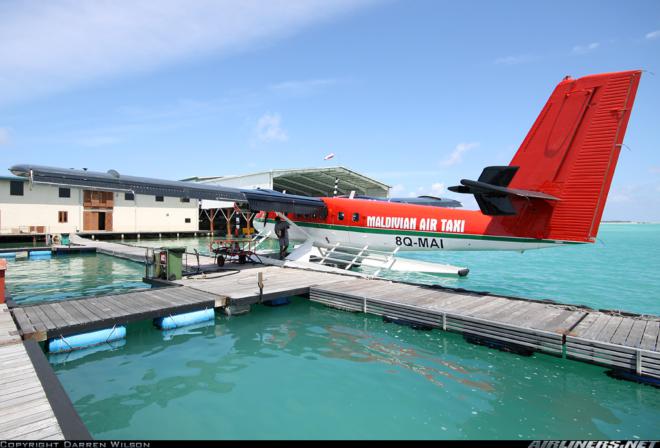}&
    \includegraphics[width=.14\textwidth]{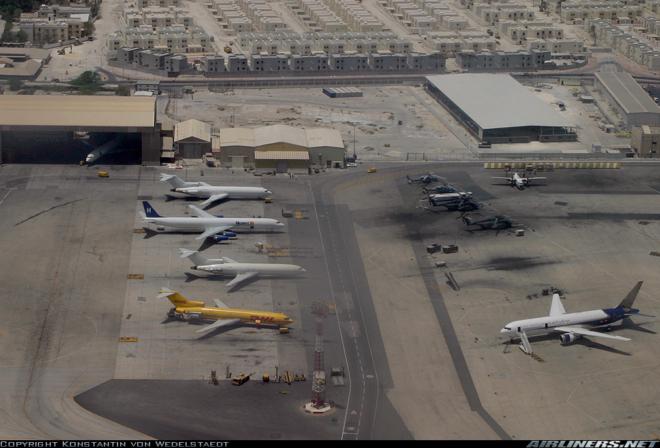}&
    \includegraphics[width=.14\textwidth]{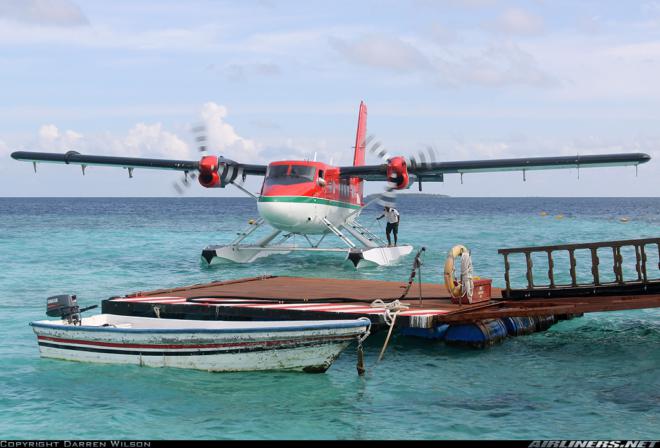}&
    \includegraphics[width=.14\textwidth]{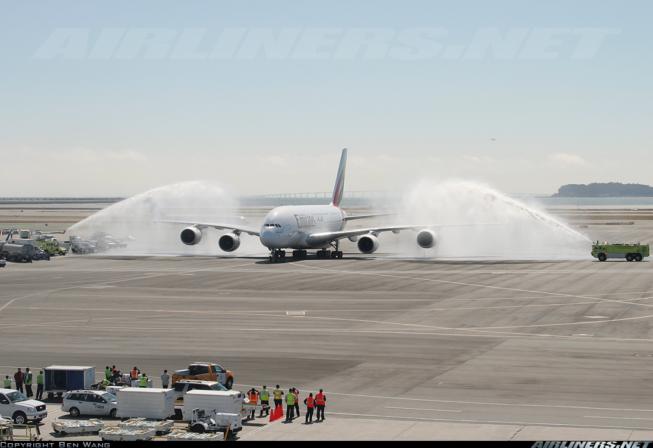}&
    \includegraphics[width=.14\textwidth]{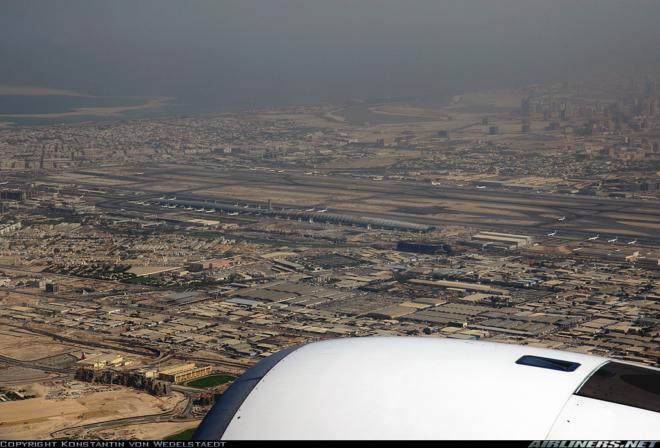}&
    \includegraphics[width=.14\textwidth]{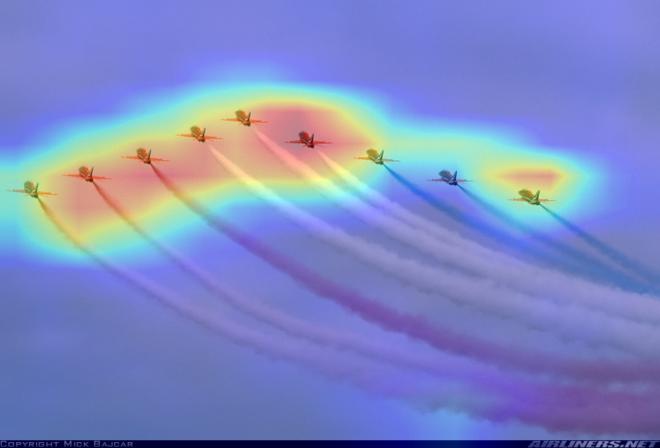}\\

    \begin{sideways} \scriptsize{Baseline G-CAM} \end{sideways}    
    \includegraphics[width=.14\textwidth]{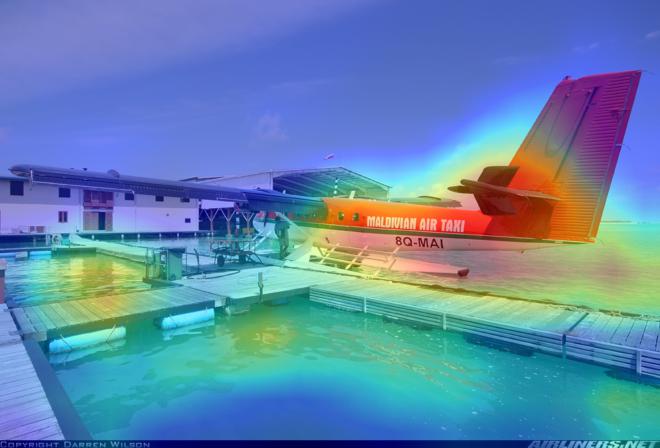}&
    \includegraphics[width=.14\textwidth]{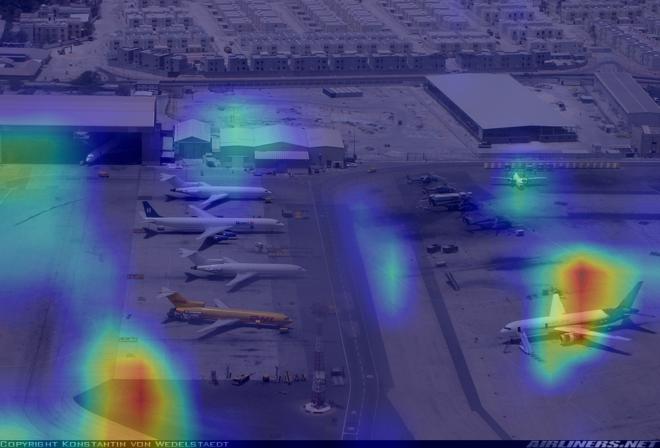}&
    \includegraphics[width=.14\textwidth]{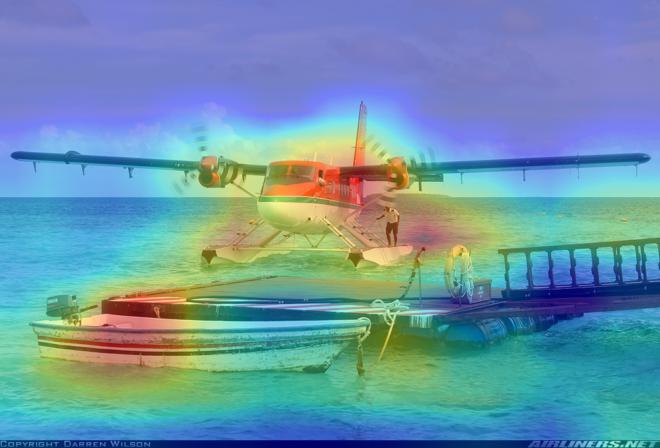}&
    \includegraphics[width=.14\textwidth]{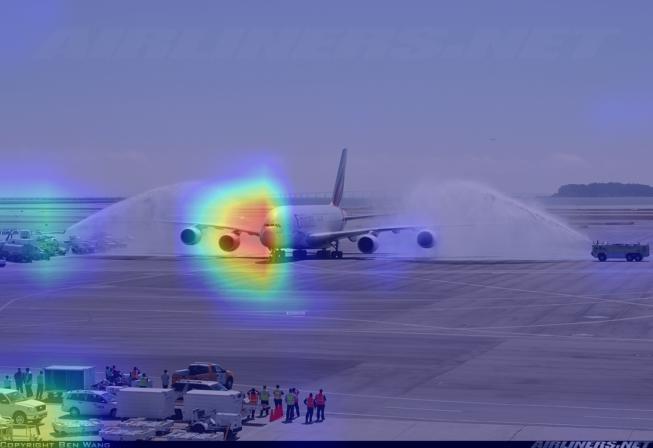}&
    \includegraphics[width=.14\textwidth]{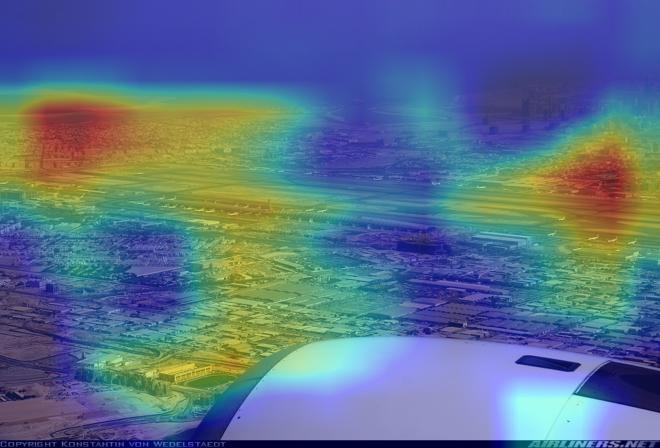}&
    \includegraphics[width=.14\textwidth]{figures/aircrafts/Hawk_T1_fail_sample/baseline.jpg}\\

    \begin{sideways} \hspace{0.125em} \scriptsize{Ours G-CAM} \end{sideways}    
    \includegraphics[width=.14\textwidth]{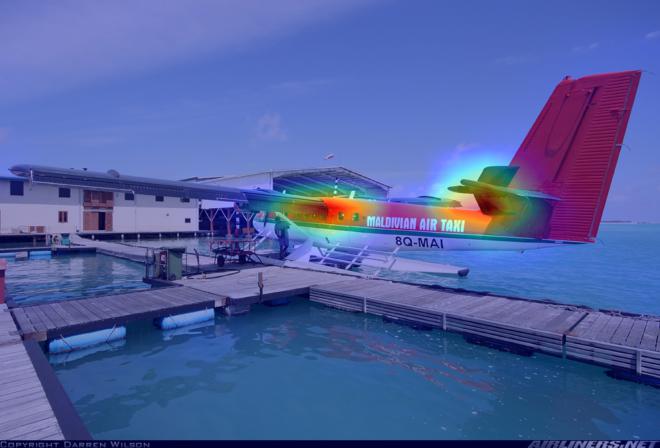}&
    \includegraphics[width=.14\textwidth]{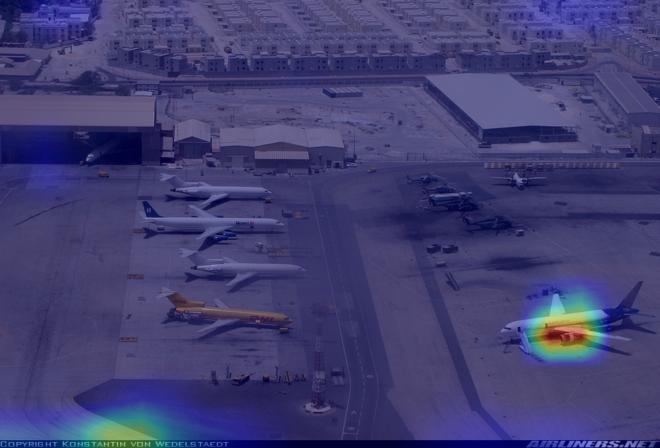}&
    \includegraphics[width=.14\textwidth]{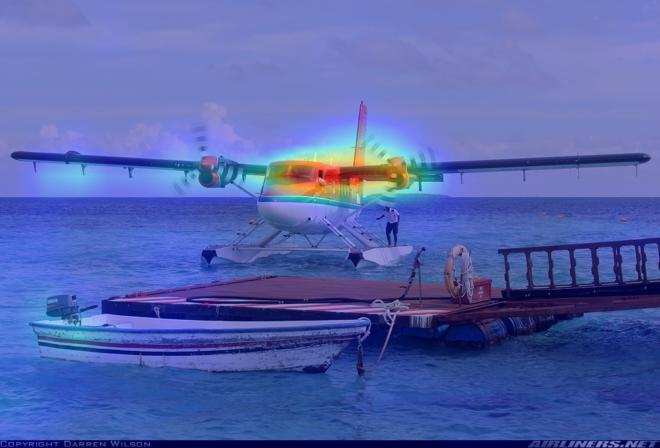}&
    \includegraphics[width=.14\textwidth]{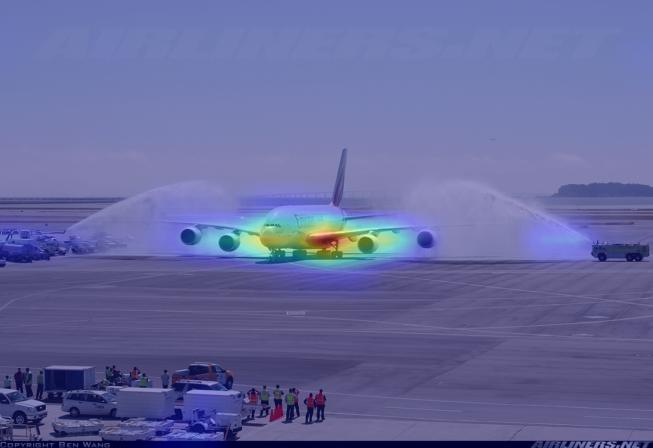}&
    \includegraphics[width=.14\textwidth]{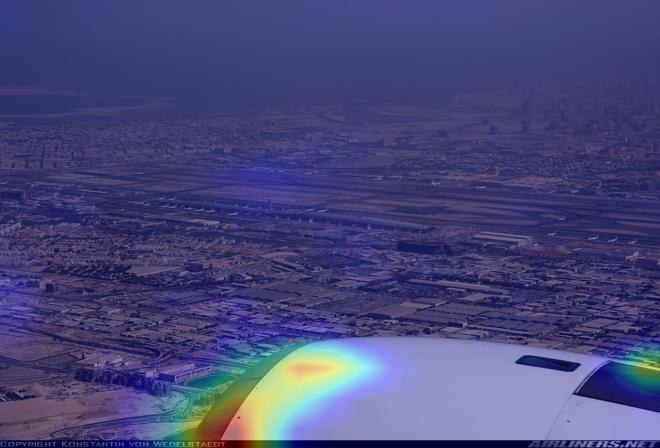}&
    \includegraphics[width=.14\textwidth]{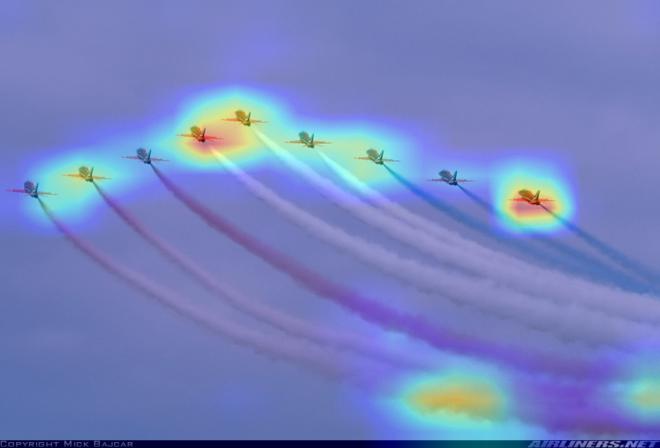}\\

\hline
\end{tabular}
\caption{Additional Grad-CAM visualization results on the FGVC Aircraft validation set using ResNet50. While our method correctly highlights most discriminative part of the aircraft in the first and the third column, the baseline incorrectly highlights the water along with the aircraft. Note that the last column shows a failure case for our model which incorrectly highlights the smoke trajectory of the aircraft.   }
\label{fig:aircrafts_gradcam_r50_vis1}
\end{figure}

\begin{figure}
\setlength{\linewidth}{\textwidth}
\setlength{\hsize}{\textwidth}
\centering
\scriptsize
\begin{tabular}{|c|c|c|c|c|c||c|}
\hline
 \small{Shoe Store} & \small{Knot} & \small{Tench} & \small{Water Hen} & \small{Albatross} & \small{Airship} & \small{Minivan}\\
 & & & & & & \\
\hline
    \begin{sideways} \quad \quad \scriptsize{Original} \end{sideways}    
    \includegraphics[width=.12\textwidth]{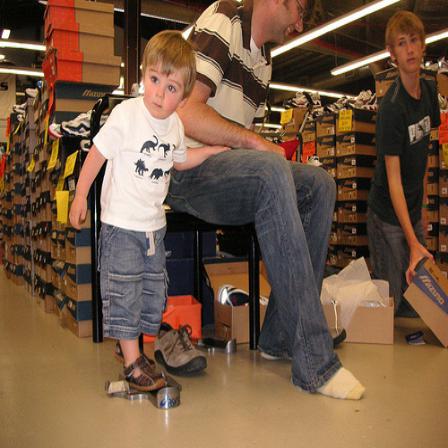}&
    \includegraphics[width=.12\textwidth]{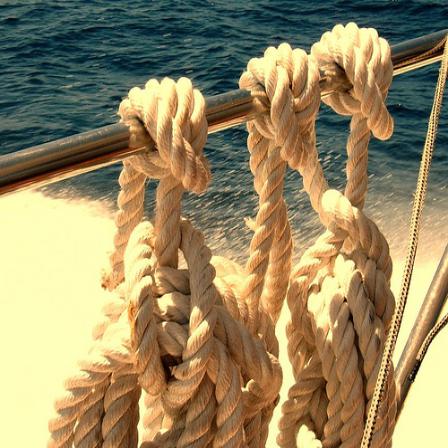}&
    \includegraphics[width=.12\textwidth]{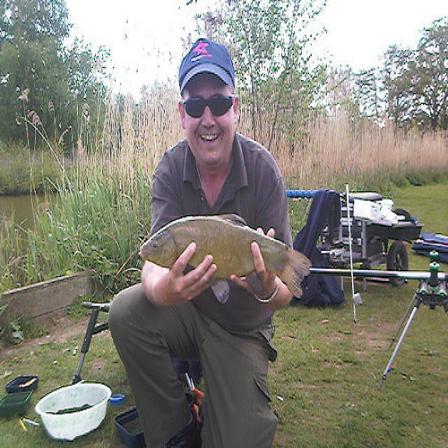}&
    \includegraphics[width=.12\textwidth]{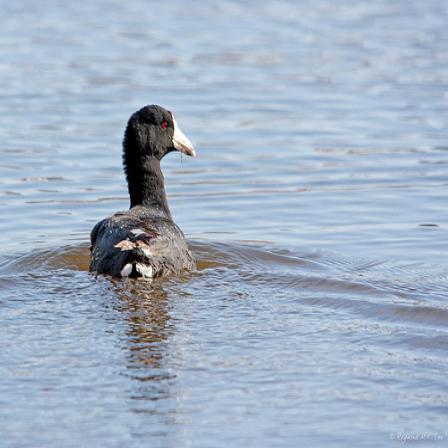}&
    \includegraphics[width=.12\textwidth]{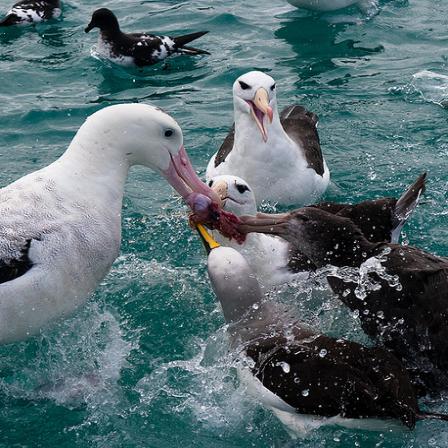}&
    \includegraphics[width=.12\textwidth]{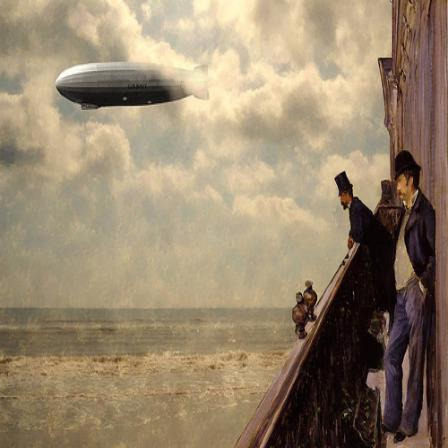}&
    \includegraphics[width=.12\textwidth]{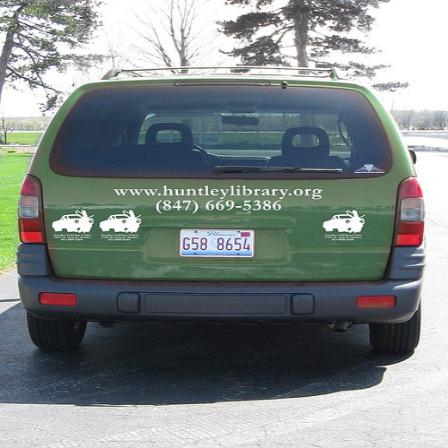}\\

    \begin{sideways} \hspace{0.25em} \scriptsize{Baseline G-CAM} \end{sideways}    
    \includegraphics[width=.12\textwidth]{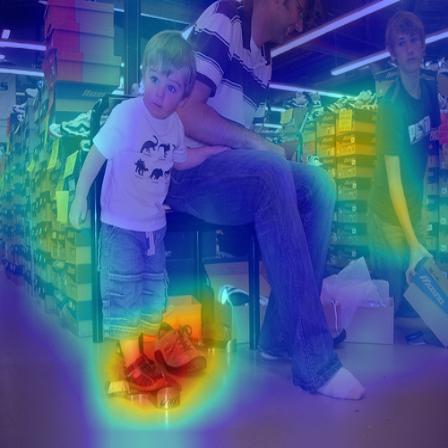}&
    \includegraphics[width=.12\textwidth]{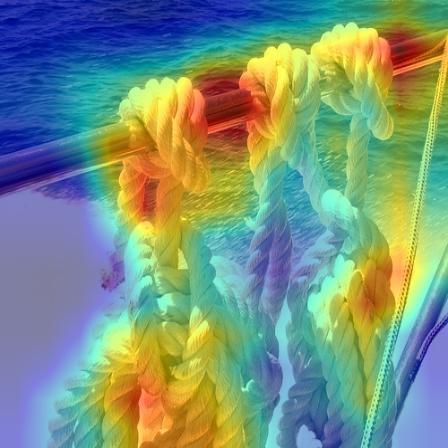}&
    \includegraphics[width=.12\textwidth]{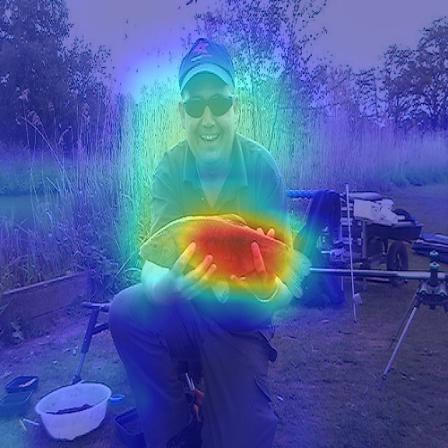}&
    \includegraphics[width=.12\textwidth]{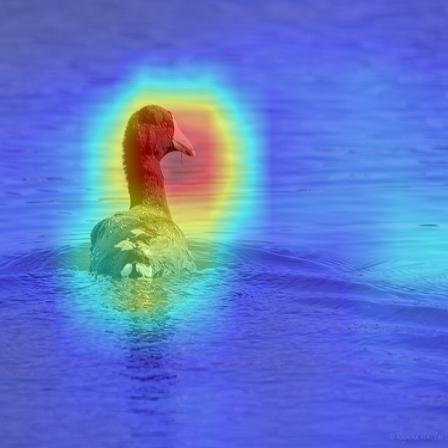}&
    \includegraphics[width=.12\textwidth]{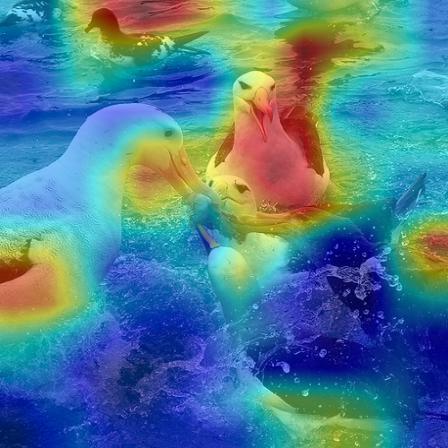}&
    \includegraphics[width=.12\textwidth]{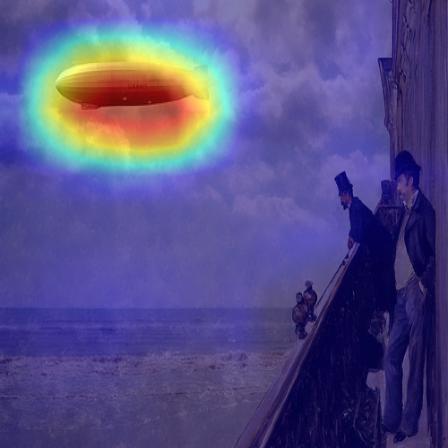}&
    \includegraphics[width=.12\textwidth]{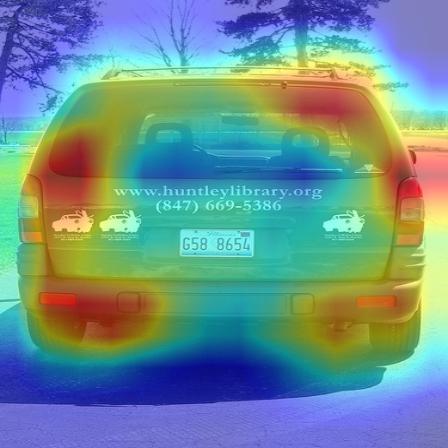}\\ 

    \begin{sideways} \quad \scriptsize{Ours G-CAM} \end{sideways}    
    \includegraphics[width=.12\textwidth]{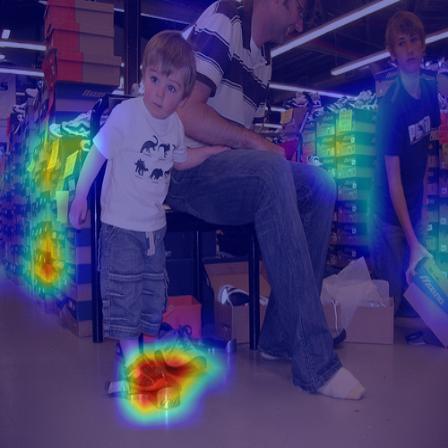}&
    \includegraphics[width=.12\textwidth]{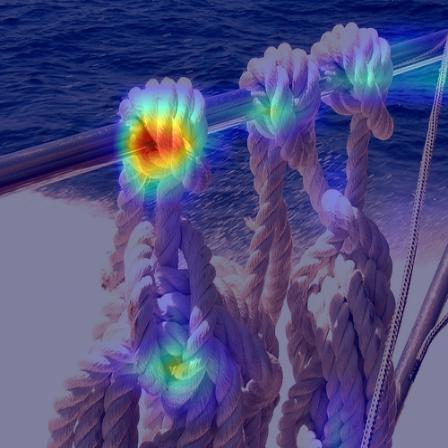}&
    \includegraphics[width=.12\textwidth]{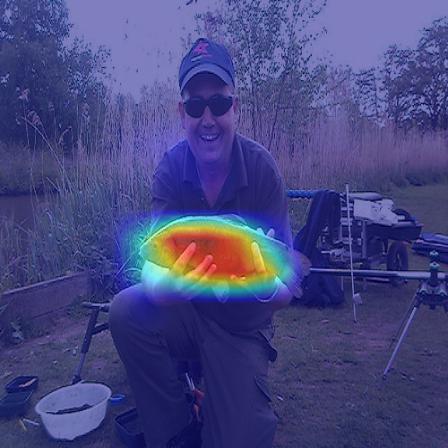}&
    \includegraphics[width=.12\textwidth]{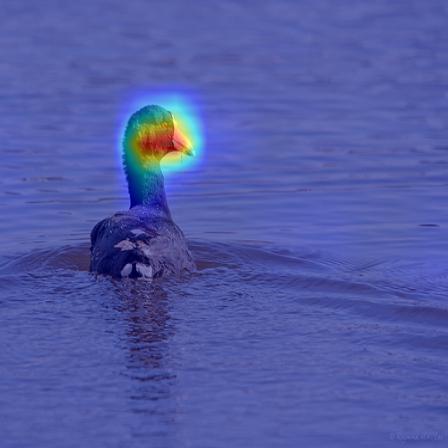}&
    \includegraphics[width=.12\textwidth]{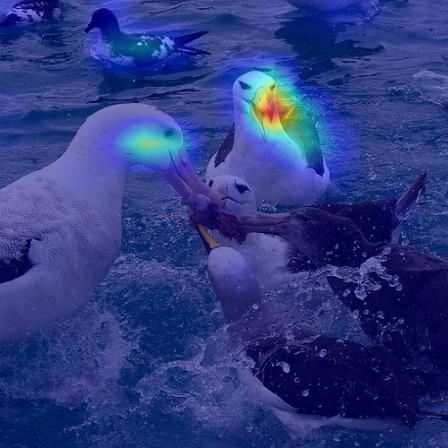}&
    \includegraphics[width=.12\textwidth]{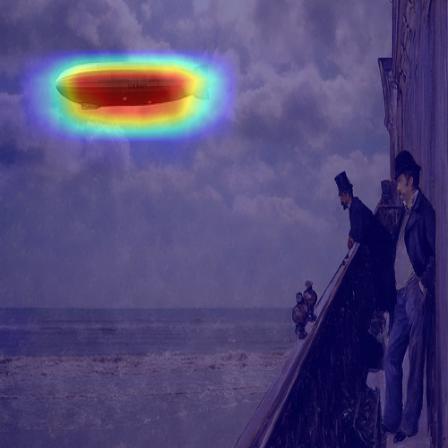}&
    \includegraphics[width=.12\textwidth]{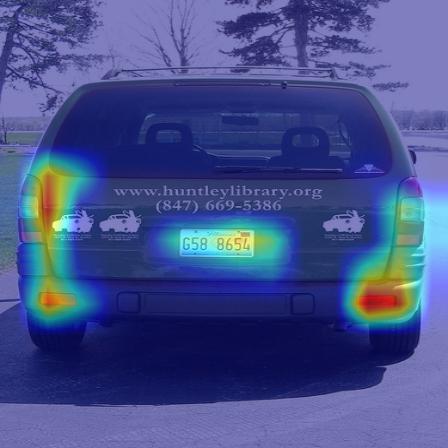}\\

\hline
\end{tabular}
\caption{Additional Grad-CAM visualization results on the ImageNet validation set using ResNet50. Our model is able to improve upon the baseline by not relying on background pixels and instead focusing on the most discriminative regions of the object. In the 3rd column, Grad-CAM is computed for the category ``Tench''and we see that the baseline incorrectly highlights the person along with the fish whereas our model correctly highlights the fish. The last column shows a failure case where our model incorrectly highlights the license plate along with the other regions of the minivan.}
\label{fig:imagenet_gradcam_r50_vis2}
\end{figure}

\begin{figure} [H]
    \centering
    \begin{tabular}{|c|c|c|c|}
    \hline
        Original & Baseline & CGC w/o neg & CGC\\
        % & & \scriptsize{negatives} & \\
        \hline
        \begin{sideways} \quad \quad Parking meter \end{sideways} 
        \includegraphics[width=.2\textwidth]{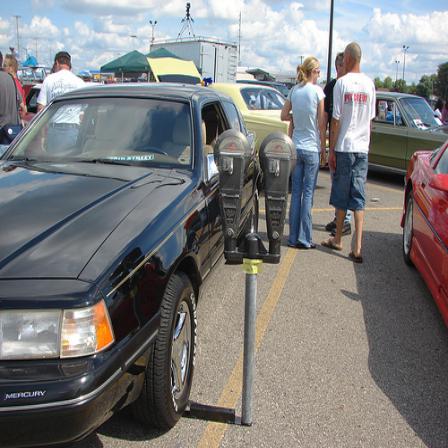}&
        \includegraphics[width=.2\textwidth]{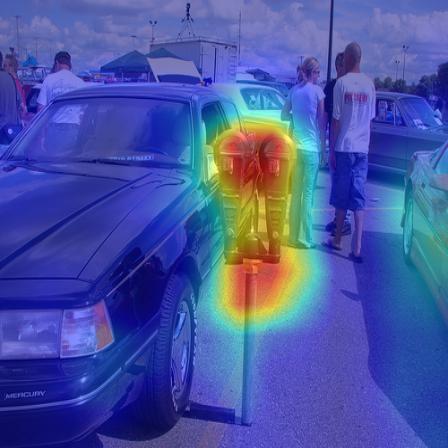}&
        \includegraphics[width=.2\textwidth]{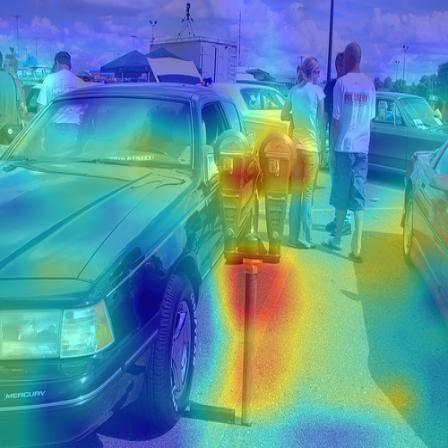}&
        \includegraphics[width=.2\textwidth]{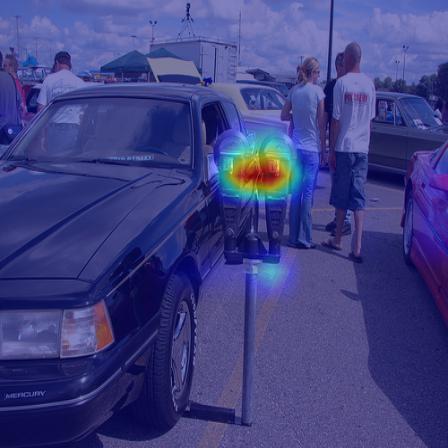}\\
        \hline
        \begin{sideways} \quad \quad Volley ball \end{sideways} 
        \includegraphics[width=.2\textwidth]{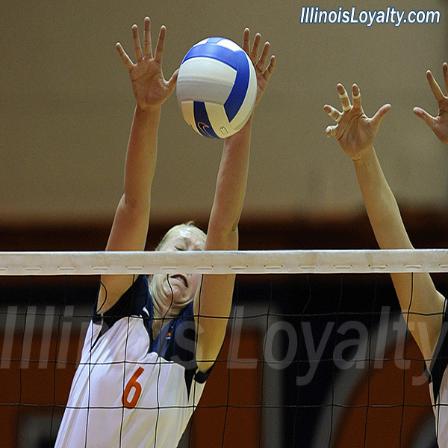}&
        \includegraphics[width=.2\textwidth]{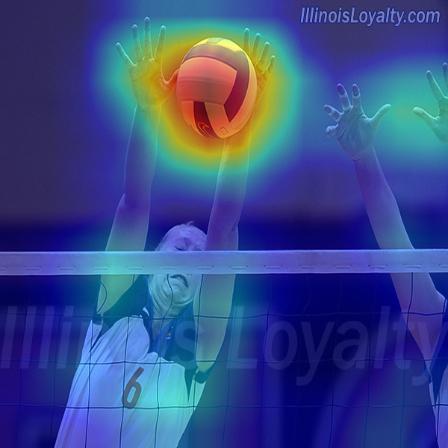}&
        \includegraphics[width=.2\textwidth]{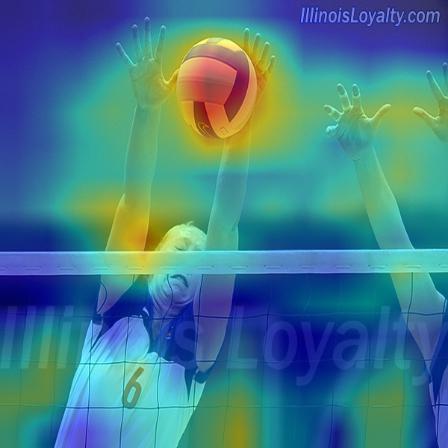}&
        \includegraphics[width=.2\textwidth]{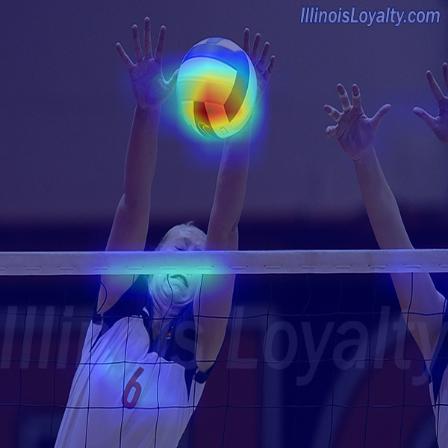}\\
        \hline
    \end{tabular}
    \caption{Qualitative results for Table 5 in the main paper. The Grad-CAM heatmap for the model trained with CGC loss, but without the negative examples results in a uniform heatmap spread across the image (column 3). In the second row, the baseline model relies on the arms of the player for classifying the image as 'Volleyball', whereas our method is able to reduce this spurious correlation.}
    \label{fig:fig_cgc_wo_negs}
\end{figure}

\begin{table}[H]
    \begin{center}
        % \begin{tabular}{|l|c|c|}
        \scalebox{1.0}{
        \begin{tabular}{|l|c|c|}
        \hline
        \multirow{2}{*}{Model} &  \multicolumn{2}{|  c  |}{ResNet50 ImageNet-100} \\ \cline{2-3}
             & Top-1 Acc (\%) & CH (\%) \\\hline
             Baseline  & \textbf{86.40 } & 53.60 \\
             CGC & 84.04 & \textbf{72.32}\\
             CGC w/o neg & 81.94 & 38.46\\
        \hline
        \end{tabular}
        }
    \end{center}
    \vspace{-0.15in}
    \caption{Results similar to Table 5 of the main paper with ResNet50 on ImageNet-100 (subset of ImageNet with 100 classes). The low CH for CGC w/o negatives shows that this method could result in heatmaps diffused across the image. The model trained without the negative heatmaps as part of $L_{CGC}$ loss results in a very low CH, thus confirming our hypothesis that the lack of negative heatmaps would result in a model learning a trivial solution of generating heatmaps diffused throughout the image.}
    \label{tab:class_ch_cgc_without_contrast_supp}
\end{table}

\section{License for assets}
% For each of the dataset and code assets we use for our experiments, we list the license below:\\
\noindent We list the license for each of the dataset and code assets used for our experiments.\\

\noindent \textbf{ImageNet:} We have been granted access to the ImageNet \cite{deng2009imagenet} dataset for non-commercial research/educational purposes and we abide by the terms of the license of this dataset.\\

\noindent \textbf{CUB-200:} This dataset was introduced in \cite{welinder2010caltech} and we use the images and the accompanying annotations for non-commercial/education research purposes only.\\

\noindent \textbf{FGVC-Aircraft:} The images in the FGVC-Aircraft \cite{maji13fine-grained} dataset has been made available exclusively for non-commercial research/educational purposes and as such we only use this dataset for non-commercial research/educational purposes.\\

\noindent \textbf{Stanford Cars-196:} This dataset \cite{KrauseStarkDengFei-Fei_3DRR2013} has been made available for non-commercial research purposes only and we abide by the terms of the license of this dataset.\\

\noindent \textbf{VGG Flowers-102:} The images and annotations in the VGG Flowers-102 \cite{Nilsback08} dataset are released under the GNU General Public License, version 2.\\

\noindent \textbf{TorchRay:} This framework was introduced in \cite{fong2019understanding} and is licensed under CC-BY-NC. We use this framework for evaluating the explanation heatmaps using the Content Heatmap (CH) metric. \\

\noindent \textbf{Insertion AUC metric:} This metric was introduced in \cite{Petsiuk2018rise} and the accompanying code is released under the MIT license.

\end{document}